\documentclass[aip,cha,12pt,amsmath,amssymb,author-year]{revtex4-1}

\usepackage{graphicx}
\usepackage[utf8]{inputenc}
\usepackage[T1]{fontenc}
\usepackage{lmodern}
\usepackage{xcolor}
\usepackage{mathtools}
\usepackage{amsthm}
\usepackage{tikz}
\usetikzlibrary{arrows.meta}
\usepackage{booktabs}
\usepackage{makecell}
\usepackage{placeins}
\usepackage[hidelinks]{hyperref}

\usepackage{etoolbox}

\makeatletter
\def\@email#1#2{%
 \endgroup
 \patchcmd{\titleblock@produce}
  {\frontmatter@RRAPformat}
  {\frontmatter@RRAPformat{\produce@RRAP{*#1\href{mailto:#2}{#2}}}\frontmatter@RRAPformat}
  {}{}
}%
\makeatother

\newtheorem{theorem}{Theorem}
\newtheorem{lemma}[theorem]{Lemma}

\newtheorem{corollary}[theorem]{Corollary}
\theoremstyle{definition}
\newtheorem{definition}[theorem]{Definition}
\theoremstyle{remark}
\newtheorem{remark}[theorem]{Remark}

\newcommand{\RR}{\mathbb{R}}

\newcommand{\norm}[1]{\left\|#1\right\|}
\newcommand{\abs}[1]{\left|#1\right|}
\newcommand{\defeq}{\coloneqq}
\newcommand{\set}[1]{\left\{#1\right\}}
\newcommand{\fh}[1]{\hat f_{#1}}
\DeclareMathOperator{\Fix}{Fix}

\definecolor{pathblue}{HTML}{0072B2}
\definecolor{pathorange}{HTML}{E69F00}
\definecolor{pathgray}{HTML}{666666}
\definecolor{pathgreen}{HTML}{009E73}

\begin{document}

\title{Error Bounds and Periodic-Point Approximation in Autonomous Echo State Networks}

\author{Taheer Jooma Abbajee\textsuperscript{*}}
\email{taheerjooma@gmail.com}
\affiliation{Department of Mathematics and Applied Mathematics, University of Johannesburg, Johannesburg, South Africa}

\author{Keegan Doig Anderson}
\affiliation{Department of Mathematics and Applied Mathematics, University of Johannesburg, Johannesburg, South Africa}
\affiliation{National Institute for Theoretical and Computational Sciences, Stellenbosch University, Stellenbosch, South Africa}

\author{Muaaz Bhamjee}
\affiliation{Clean Energy Research Group, Department of Mechanical and Aeronautical Engineering, University of Pretoria, Pretoria, South Africa}

\author{Maria Vivien Visaya}
\affiliation{Department of Mathematics and Applied Mathematics, University of Johannesburg, Johannesburg, South Africa}
\affiliation{National Institute for Theoretical and Computational Sciences, Stellenbosch University, Stellenbosch, South Africa}

\date{\today}

\begin{abstract}
A reservoir computer trained only to predict one step of a dynamical system can be run autonomously by feeding its own prediction back as input. Its multistep predictions then approximate iterates of the map, although the reservoir is never trained on them. We establish explicit bounds on what these predictions inherit from the iterated map. If the reservoir has just received a periodic point of the map, its autonomous prediction one period later returns within a computable distance of that point. A global bound on the multistep prediction error depends on the Lipschitz constant of the map, a sharper local bound depends on the Jacobians along the orbit, and a residual bound shows that a periodic point of the autonomous reservoir is an approximate periodic point of the map. A Lyapunov-rate heuristic defines a practical prediction horizon. We validate the framework on the logistic map, the H\'enon map, and a time-discretized Lorenz system, with 50 reservoirs per system. No proven bound is violated in 3.5 million admissible checks, whereas a negative control with reduced constants fails in 91\% of its checks. The local bound remains informative for 1.8 to 3.3 times as many steps as the median global bound. For the two maps, the periodic orbits of the autonomous reservoir match every true periodic point of period at most four, with none spurious. For the Lorenz system, reservoir orbits of period 16 lie within $8\times10^{-4}$ of the shortest periodic orbit of the discretized flow.
\end{abstract}

\maketitle

\begin{quotation}
Reservoir computers are efficient neural networks that have become a popular tool for forecasting chaotic systems. A reservoir is trained only to predict one step ahead and is then run on its own, feeding each prediction back in as its next input. It is natural to ask how much such a self-driven reservoir really knows about the underlying system beyond that single step: how far its multistep forecasts can be trusted, and whether it has captured the periodic orbits, the repeating patterns that form the skeleton of chaotic motion. We prove explicit error bounds that answer these questions and that can be computed whenever the true system is known. Numerical tests on three standard chaotic systems show that the bounds are never violated, that a sharper trajectory-dependent bound stays informative several times longer than the worst-case bound, and that the reservoir recovers periodic orbits it was never explicitly trained on.
\end{quotation}

\section{Introduction}
\label{sec:intro}
Chaotic dynamical systems are deterministic, yet their sensitive dependence on initial conditions renders long-term forecasting from observed time series intrinsically difficult. Reservoir computing (RC) has established itself as an effective data-driven approach to this problem. A fixed, randomly generated recurrent network, the reservoir, is driven by an input time series, and only a linear readout is trained, typically by least squares \citep{jaeger2001,maass2002,jaeger2004}. Reservoir computers predict chaotic systems over short and medium horizons \citep{jaeger2004,pathak2018,platt2022}, and a trained reservoir operated autonomously can reproduce long-term properties such as attractor geometry, ergodic statistics and Lyapunov exponents \citep{lu2018,pathak2017,krishnagopal2019,hartjd2024}. These observations are supported by a growing body of theory, comprising universal approximation results \citep{grigoryevaortega2018}, embedding and approximation theorems \citep{hart2020,grigoryeva2021,grigoryeva2023}, a description of the reservoir--system relation through generalized synchronization \citep{lymburn2019}, and approximations of the Koopman operator constructed from reservoir states \citep{gulina2021}. In particular, \citet{hara2024} analyze RC from the viewpoint of learning dynamics and show that a trained reservoir can learn information about the dynamics that is not contained in the training data.

These contributions concern either the one-step map or long-run statistics. Fixed points, periodic orbits and Lyapunov exponents of the autonomous reservoir have also been examined as properties of a data-driven model \citep{kobayashi2021}. More recently, \citet{ChenInubushi2026} linearly combine readouts trained at two parameter values of the same reservoir, under a generalized synchronization assumption, bound the error of the combined readout and the resulting one-step conjugacy residual, in order to extrapolate learned dynamics across bifurcations. Reservoir computing has also been combined with adaptive delayed feedback control to detect unstable periodic orbits from time series alone \citep{zhu2019}. What has been less developed is a quantitative guarantee: how closely the $n$-step autonomous predictions reproduce the $n$-fold map $f^n$ and its periodic orbits, which constitute the skeleton of a chaotic attractor \citep{auerbach1987}, with explicit error control that can be evaluated when $f$ is known.

We study this question for a reservoir trained only on the one-step map $f$, where $x_{t+1}=f(x_t)$. Let $x_t$ denote the current state and $r_t$ the reservoir state that has just received $x_t$. Feeding the readout back as the next input defines the autonomous reservoir map $G$ and, for each $n\ge1$, the $n$-step predictor $\fh{n}$ of Definition \ref{def:predictor}, with $\fh{1}(x_t)=W_{\text{out}}r_t$. The predictor $\fh{n}(x_t)$ estimates $f^n(x_t)$ although the reservoir is never trained on $f^n$. Since $r_t$ depends on the driving history and not on $x_t$ alone, $\fh{n}(x_t)$ is a function of the pair $(x_t,r_t)$, and it is not the $n$-th power of a fixed map. Three questions then arise that one-step accuracy alone does not settle. First, if $f^n(x^*)=x^*$, does the $n$-step prediction started at $x^*$ return close to $x^*$, and can approximate periodic points of $f$ be located from the autonomous reservoir? Second, how does the error $E_n=\norm{\fh{n}(x_t)-f^n(x_t)}$ accumulate with the number of steps $n$, and can its growth be controlled analytically? Third, up to which $n$ does prediction against a specific orbit remain informative?

The contributions are twofold: a quantitative connection between one-step approximation errors, autonomous iteration and periodic-orbit residuals, and a distinction between global and trajectory-dependent local error control. They are stated as three theorems and their corollaries. If the reservoir has just received a point $x^*$ with $f^n(x^*)=x^*$, then $\fh{n}(x^*)$ lies within $B_L(n)\varepsilon$ of $x^*$, where $\varepsilon$ bounds the one-step error along the orbit and $B_L(n)$ depends on $n$ and on the Lipschitz constant $L$ of $f$; conversely, an approximate fixed point of $G^n$ yields an approximate point of period $n$ of $f$ (Theorem \ref{thm:fixed_point_preservation}). Since $\Fix(f^n)$ is the union of the periodic orbits of $f$ whose prime period divides $n$, this permits locating approximate periodic points by root finding on $G^n$ alone, with certification requiring $f$ (Corollary \ref{cor:periodic_detection}). The converse, namely that $G$ possesses a periodic orbit near every true one, is not proved and is examined empirically. The error $E_n$ satisfies a global geometric bound in $L$ (Theorem \ref{thm:error_propagation}) and a sharper local bound in the Jacobian norms realized along the orbit (Theorem \ref{thm:local_jacobian}); the latter is motivated by finite-time Lyapunov growth, although this connection is heuristic and does not follow from the theorem. Finally, comparing a Lyapunov-rate estimate of the error with the diameter of the attractor defines a practical prediction horizon $n_{\max}$ (Definition \ref{def:prediction_horizon}), which is a calibrated heuristic rather than a proven cutoff.

All bounds are conditional: they hold whenever the true and autonomous orbits remain in a compact convex domain $K$ and the one-step errors along them are bounded by $\varepsilon$. The quantity $\varepsilon$ is not predicted from the training data but measured, and the bounds are then verified against the observed errors. Section \ref{sec:numerical} reports this verification on the logistic map, the two-dimensional H\'enon map and a time-discretized Lorenz system, with 50 reservoir realizations per system and a negative control demonstrating that the checks are capable of failing.

\section{Reservoir Computing Model}
\label{sec:rcmodel}
\begin{figure}
		\centering
		\resizebox{\columnwidth}{!}{%
			\begin{tikzpicture}[
				node/.style={circle, draw, line width=1.25pt, minimum size=0.56cm, inner sep=0pt, fill=white},
				inedge/.style={pathblue, densely dashed, -{Triangle[length=1.9mm,width=2.0mm]}, line width=1.25pt},
				outedge/.style={pathorange, -{Triangle[length=1.9mm,width=2.0mm]}, line width=1.25pt},
				resedge/.style={pathgray, densely dashed, -{Triangle[length=1.4mm,width=1.5mm]}, line width=1.0pt},
				every node/.style={font=\fontsize{24}{28}\selectfont},
				]

                \draw[line width=3.6pt, rounded corners=11pt] (3.6,0.8) rectangle (10.7,6.15);

				\node[node] (A) at (4.50,4.75) {};
				\node[node] (B) at (6.30,5.20) {};
				\node[node] (C) at (8.90,5.30) {};
				\node[node] (D) at (7.20,3.80) {};
				\node[node] (E) at (9.55,3.90) {};
				\node[node] (F) at (5.30,3.50) {};
                \node[node] (G) at (7.00,2.20) {};
				\node[node] (H) at (8.50,2.60) {};
				\node[node] (I) at (4.85,1.75) {};

				\draw[resedge] (A) to[out=85,  in=190] (B);
				\draw[resedge] (C) to[out=185, in=355] (B);
				\draw[resedge] (A) to[out=20,  in=180] (D);
				\draw[resedge] (E) to[out=185, in=5]   (D);
				\draw[resedge] (E) to[out=100, in=310] (C);
				\draw[resedge] (B) to[out=-50, in=105] (H);
				\draw[resedge] (D) to[out=-100,in=110] (G);
				\draw[resedge] (H) to[out=195, in=15]  (G);
				\draw[resedge] (F) to[out=100, in=270] (A);
				\draw[resedge] (F) to[out=-160,in=155] (I);
				\draw[resedge] (I) to[out=-5,  in=205] (G);
				\draw[resedge] (G) to[out=135, in=330] (F);

				\coordinate (xin) at (1.8,3.5);
				\draw[inedge] (xin) -- (3.3,4.85);
				\draw[inedge] (xin) -- (3.3,3.5);
				\draw[inedge] (xin) -- (3.3,2.1);

				\draw[outedge] (11.0,4.9) -- (12.51,3.78);
                \draw[outedge] (11.0,3.5) -- (12.51,3.47);
				\draw[outedge] (11.0,2.1) -- (12.51,3.15);

				\node at (0.8,6.6)  {Input};
				\node at (7.15,6.6) {Reservoir};
				\node at (13.41,6.6) {Output};

				\node[pathblue] at (0.7,3.5)   {$x_{t}$};
				\node[pathblue] at (1.5,2.05)  {$W_{\text{in}}$};

				\node[pathorange] at (13.36,3.5) {$y_{t}$};
				\node[pathorange] at (12.86,2.05){$W_{\text{out}}$};

				\node at (7.15,0.3)  {$r_{t}$};
				\node[pathgray] at (9.45,1.4)  {$W_{\text{res}}$};

                \draw[pathgreen, line width=6pt, line join=miter, line cap=butt]
					(14.05,3.5) -- (14.85,3.5) -- (14.85,-0.8)
					-- node[midway, fill=white, text=pathgreen, inner sep=6pt, font=\fontsize{21}{25}\selectfont\bfseries] {Autonomous operation} (-0.95,-0.8)
					-- (-0.95,3.5) -- (-0.35,3.5);
				\fill[pathgreen] (-0.35,3.88) -- (0.1,3.5) -- (-0.35,3.12) -- cycle;
            \end{tikzpicture}
		}
		\caption{Schematic of the reservoir computing architecture. Dashed connections denote weights fixed at initialization, with blue and gray indicating the input weights $W_{\text{in}}$ and recurrent reservoir weights $W_{\text{res}}$, respectively. The solid orange connection denotes the readout weights $W_{\text{out}}$, which is the only trained component. The thick green arrow indicates autonomous operation, in which the output is fed back as the input.}
		\label{fig:rcmodel}
\end{figure}
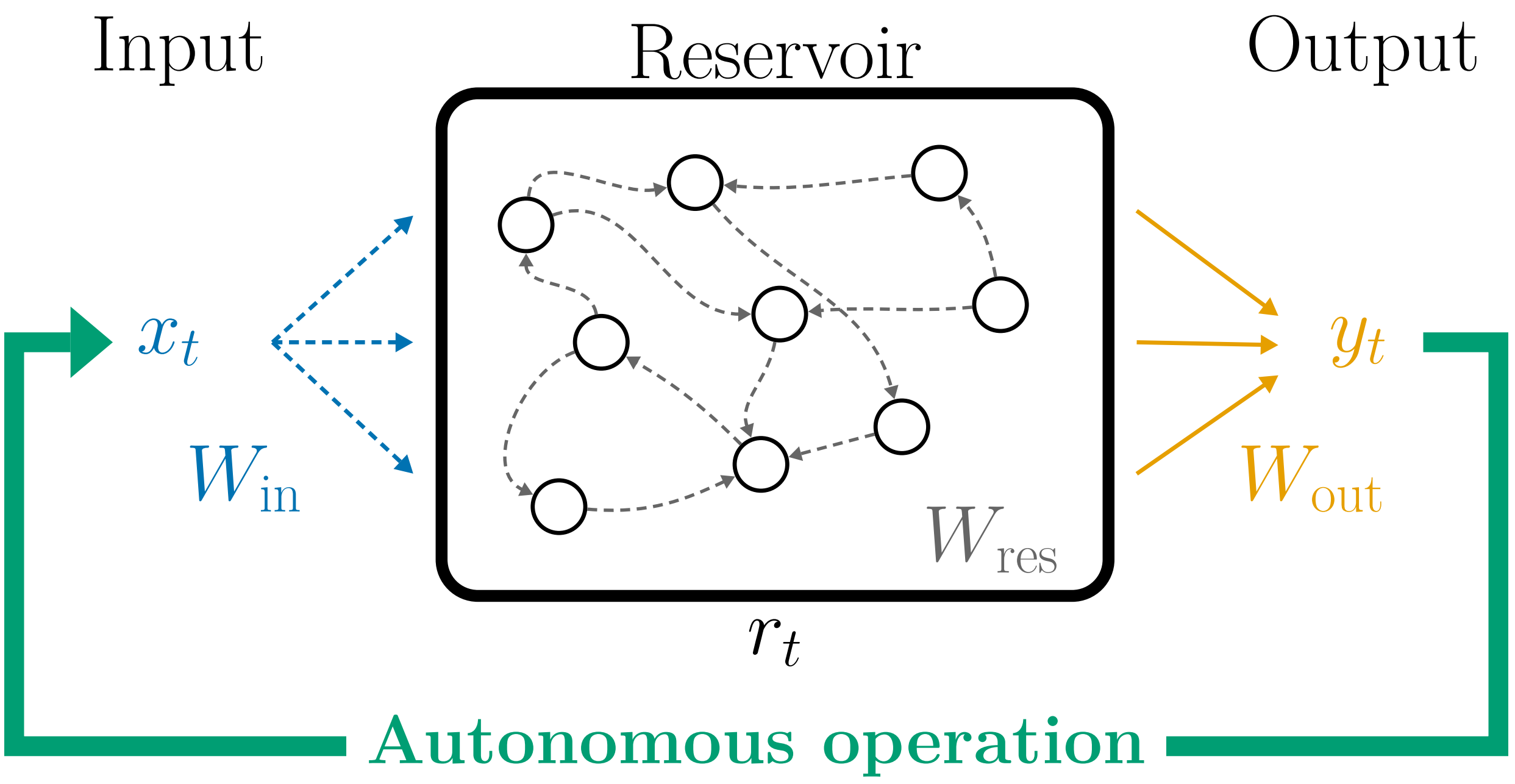

The reservoir computing architecture considered throughout this paper is illustrated in Figure \ref{fig:rcmodel}. For an input--output space of dimension $d_x$ and a reservoir of dimension $d_r$, the reservoir state $r_t\in\RR^{d_r}$ is the state that has just received the input $x_t$, and evolves according to
	\begin{equation}
		\label{eq:reservoir-update}
		r_{t} = F(r_{t-1},x_t)= \tanh\big(W_{\text{res}}r_{t-1}+W_{\text{in}}x_t+b\big),\qquad t\ge1,
	\end{equation}
where $F:\RR^{d_r}\times\RR^{d_x}\to\RR^{d_r}$ is the reservoir update map, $x_t\in\RR^{d_x}$ is the driving input, $W_{\text{in}}\in\RR^{d_r\times d_x}$ is the input weight matrix, $W_{\text{res}}\in\RR^{d_r\times d_r}$ is the recurrent reservoir weight matrix, and $b\in\RR^{d_r}$ is a constant bias. The hyperbolic tangent is applied element-wise.

The entries of $W_{\text{in}}$ are drawn uniformly from $[-\sigma,\sigma]$. The reservoir matrix $W_{\text{res}}$ is the symmetric weighted adjacency matrix of an Erd\H{o}s--R\'enyi graph \citep{erdds1959random} on $d_r$ nodes without self-loops: each of the $d_r(d_r-1)/2$ possible edges is absent independently with probability $\tau\in[0,1)$ (the sparsity), and the weight of each present edge is drawn uniformly from $[-1,1]$. The matrix is then rescaled so that its spectral radius equals the prescribed value $\rho$; if the sampled graph has no edges, the rescaling is undefined and the graph is redrawn (this did not occur for any seed used). The bias is given by $b=\beta 1_{d_r}$, where $\beta \in (-1, 1)$ and $1_{d_r}\in\mathbb{R}^{d_r}$ denotes the vector of all ones of dimension $d_r$. Once initialized, $W_{\text{in}}$, $W_{\text{res}}$, and $b$ are held fixed throughout training and prediction.

The reservoir state is mapped to the output through a linear readout,
	\begin{equation}
		\label{eq:readout}
		y_t=W_{\text{out}}r_t, \qquad W_{\text{out}} \in\RR^{d_x\times d_r}.
	\end{equation}
The readout is the only trainable component and is fitted by ordinary least squares, as described in Section \ref{sec:training}.

\subsection{Echo state property and synchronization}
\label{sec:gs-esp}
A reservoir has the echo state property (ESP) \citep{jaeger2001} if, for a given input sequence, trajectories started from different initial reservoir states converge to one another, so that after a transient $r_t$ is determined by the input history alone. The condition that the spectral radius of $W_{\text{res}}$ be below $1$ is commonly used as a heuristic proxy for the ESP, although it is neither necessary nor sufficient in general \citep{yildiz2012,manjunathjaeger2013}; the stronger, global condition is on the spectral norm, $\norm{W_{\text{res}}}_2<1$ \citep{jaeger2001, jaeger2004}, and activation-specific contraction criteria are given by \citet{singh2025}. For the symmetric $W_{\text{res}}$ used here, the spectral radius equals the spectral norm, and both equal $\rho<1$ by construction, so the two conditions coincide. Since $\tanh$ is $1$-Lipschitz, this condition ensures that the reservoir update is contractive in its state for every fixed input. The ESP therefore holds for every input sequence, and the warm-up phase of Section \ref{sec:training} reduces the influence of the arbitrary initial state.

A complementary, more informative diagnostic is the largest conditional Lyapunov exponent: linearizing the driven map $r\mapsto F(r,x)$ along a trajectory generated by a fixed input sequence $\{x_t\}$ gives an infinitesimal perturbation $\xi_{t+1}=D_rF(r_t,x_{t+1})\xi_t$, propagated with the drive held fixed, whose growth rate over $T$ steps for a generic initial perturbation $\xi_0\neq0$ defines
\begin{equation}
	\label{eq:cle}
	\lambda_{\text{cond}} \defeq \limsup_{T\to\infty} \frac{1}{T} \ln\left(\frac{\norm{\xi_T}}{\norm{\xi_0}}\right).
\end{equation}
A negative $\lambda_{\text{cond}}$ indicates convergence of reservoirs started from different states, and is the diagnostic most commonly used for the ESP in recent reservoir computing work \citep{platt2022,hartjd2024,ohkubo2024}.

When the inputs are generated by a deterministic system $x_{t+1}=f(x_t)$, the reservoir is a response system driven by $f$, and the ESP is closely related to generalized synchronization (GS) \citep{rulkov1995,kocarev1996,lymburn2019,lu2018,nazerian2023}: after the transient, the reservoir state is a function of the drive, $r_t=h(x_t)$ when $f$ is invertible (with $h$ smooth under suitable conditions \citep{grigoryeva2021}), and more generally a function of the drive's past orbit. Together, these properties make the autonomous run of Section \ref{sec:training} well defined: the ESP fixes $r_t$ from the driving history, and the encoding of the drive in $r_t$ is what allows a linear readout trained on $r_t$ to approximate $f(x_t)$. Readouts built directly around the synchronization map have also been proposed \citep{ohkubo2024}.

\subsection{Training and Autonomous Prediction}
\label{sec:training}

\noindent\textit{Warm-up and training phases.}
Given an observed trajectory $x_1,x_2,\dots$ of the target map $f$, so that $x_{t+1}=f(x_t)$, we fix $1\le T_0<T_1$. The reservoir, initialized at $r_0=0$, is driven through the warm-up segment $\{x_t\}_{t=1}^{T_0}$ via \eqref{eq:reservoir-update} to eliminate the transient dependence on the arbitrary initial state; the states $r_t$ collected for $t=T_0,\dots,T_1-1$ are then paired with the targets $x_{t+1}$ (which requires data up to $x_{T_1}$) to fit $W_{\text{out}}$ by ordinary least squares.

\noindent\textit{One-step-ahead training target.}
Every reservoir state $r_t$, which has just received $x_t$, is paired with $x_{t+1}$,
\[
    r_t \longmapsto x_{t+1} = f(x_t),
\]
so that $W_{\text{out}}$ minimizes $\sum_{t=T_0}^{T_1-1}\norm{W_{\text{out}}r_t-f(x_t)}^2$ over the $T_1-T_0$ training pairs. The trained readout therefore satisfies $W_{\text{out}}r_t\approx f(x_t)$ along the training orbit.

\noindent\textit{Autonomous operation.}
	Once trained, the reservoir can be run without external input by feeding its own readout back as the next input, which gives the autonomous reservoir map
	\begin{equation}
		\label{eq:g-auto}
		G(r) \defeq F(r, W_{\text{out}}r) = \tanh\big(W_{\text{res}}r + W_{\text{in}}W_{\text{out}}r + b\big),
	\end{equation}
obtained by substituting the input $W_{\text{out}}r$ into \eqref{eq:reservoir-update}.

\begin{definition}[Autonomous $n$-step predictor]
    \label{def:predictor}
	Suppose the reservoir has been driven by the true orbit of $f$ beyond the warm-up (that is, $t\ge T_0$, so that it has forgotten its initial state, Section \ref{sec:gs-esp}), and let $x_t$ denote the current state and $r_t$ the reservoir state that has just received $x_t$. Since $r_t$ depends in general on the driving history and not on $x_t$ alone, the predictions below depend on $r_t$. The reservoir is then operated autonomously from time $t$ by setting $\hat x_t\defeq x_t$, $\hat r_t\defeq r_t$ and feeding each output back as the next input: for $k\ge0$,
	\begin{equation}
		\label{eq:autonomous-run}
		\hat x_{t+k+1}\defeq W_{\text{out}}\hat r_{t+k},\qquad \hat r_{t+k+1}\defeq F(\hat r_{t+k},\hat x_{t+k+1})=G(\hat r_{t+k}),
	\end{equation}
    which amounts to iterating the map $G$ in \eqref{eq:g-auto}, so that $\hat r_{t+k}=G^k(r_t)$ and $\hat x_{t+k+1}=W_{\text{out}}G^k(r_t)$. In particular, the trained one-step prediction $\fh{1}(x_t)=\hat x_{t+1}=W_{\text{out}}r_t$ is fed back as the input for the next state, giving $G(r_t)=F(r_t,\fh{1}(x_t))$, and the same procedure is repeated at each subsequent step. The $n$-step predictor is
	\begin{equation}
		\label{eq:predictor}
		\fh{n}(x_t)\defeq \hat x_{t+n}=W_{\text{out}}G^{n-1}(r_t), \qquad n\ge1,
	\end{equation}
    an estimate of $x_{t+n}=f^n(x_t)$. By convention, $G^0$ is the identity function. The exponent $n-1$ reflects that the readout of the reservoir state $\hat r_{t+n-1}$, reached after $n-1$ autonomous updates, already predicts time $t+n$.
\end{definition}

\section{Theoretical Framework}
\label{sec:theory}
The theoretical analysis focuses on how one-step approximation errors propagate under autonomous iteration and how this propagation affects periodic behavior. To make these questions precise, we first specify the domain, the properties of the true map, and the trained reservoir model used throughout the analysis.

\subsection{Setup and Standing Assumptions}
\label{sec:theory-setup}

\begin{itemize}
	\item \textbf{Domain.} $K\subseteq\RR^{d_x}$ is compact and convex, and $\operatorname{diam}(K)$ is its diameter.
	\item \textbf{True map.} $f$ is Lipschitz on $K$ with constant $L$: $\norm{f(u)-f(v)}\le L\norm{u-v}$ for $u,v\in K$. Where derivatives are used, $f$ is $C^1$ on an open neighborhood of $K$ and $L\defeq\sup_{K}\norm{Df}_2$, which is admissible because $K$ is convex.
	\item \textbf{Trained model.} $(W_{\text{in}},W_{\text{res}},b,W_{\text{out}})$ as in Section \ref{sec:rcmodel}, with autonomous map $G$ \eqref{eq:g-auto}.
\end{itemize}

\begin{definition}[Autonomous orbit, local error, effective error]
	\label{def:local-error}
	A pair $(x_t,r_t)\in\RR^{d_x}\times\RR^{d_r}$ at a time $t\ge T_0$ consists of a state $x_t$ and a reservoir state $r_t$ that has just received $x_t$, as in Definition \ref{def:predictor}. Its autonomous orbit is $\hat x_t\defeq x_t$ and $\hat x_{t+k}\defeq\fh{k}(x_t)$ for $k\ge1$; it is a pseudo-orbit of $f$ with local errors
	\[
		\delta_k\defeq\norm{\hat x_{t+k+1}-f(\hat x_{t+k})},\qquad k\ge0.
	\]
	For a set $\mathcal P$ of pairs and $n\ge1$, the effective error is
	\begin{equation}
		\label{eq:eps-eff}
		\varepsilon_n(\mathcal P)\defeq\sup_{(x_t,r_t)\in\mathcal P}\ \max_{0\le k<n}\delta_k.
	\end{equation}
\end{definition}

\noindent\textit{Notation: time $t$ and lead $k$.}
The index $t$ denotes absolute time and identifies the pair $(x_t,r_t)$, whereas $k$ denotes the lead, or the number of steps from the forecast origin $t$. Accordingly, the local errors $\delta_k$, prediction errors $E_k$, local constants $\ell_k$, and associated sums and products are indexed by lead. Since $f$ is autonomous, the true state at lead $k$ is $f^k(x_t)=x_{t+k}$. We therefore define the prediction error at lead $k$ by
\[
	E_k \defeq \norm{\hat{x}_{t+k}-x_{t+k}}, \qquad E_0=0.
\]

The effective error $\varepsilon_1(\mathcal P)$ reduces to the usual one-step error, since only $\delta_0$ contributes. For $n>1$, it accounts for the errors accumulated along the predicted trajectory, including errors evaluated at states that may not have been encountered during training. Thus, a reservoir can be accurate on the training orbit while still having a large effective error during autonomous prediction. The results below are stated in terms of a bound $\varepsilon$ on the local errors $\delta_k$, $0\le k<n$. If the true and autonomous orbits of every pair in $\mathcal P$ remain in $K$ up to lead $n-1$, one may take $\varepsilon=\varepsilon_n(\mathcal P)$ for every pair in $\mathcal P$.

\begin{definition}[Approximate fixed and periodic points]
	\label{def:approx-fixed-point}
	For a map $g:\RR^{d_x}\to\RR^{d_x}$ let $\Fix(g)\defeq\{x:g(x)=x\}$ denote the set of fixed points of $g$. For $\eta\ge0$ a point $x\in K$ is called an $\eta$-approximate fixed point of $g$ if $\norm{g(x)-x}\le\eta$. A point $x\in K$ is called an $\eta$-approximate $n$-periodic point of $g$ (period dividing $n$) if it is an $\eta$-approximate fixed point of $g^n$, meaning $\norm{g^n(x)-x}\le\eta$. The quantity $\norm{g^n(x)-x}$ is referred to as the residual of $x$.
\end{definition}

\subsection{Lipschitz Control of Autonomous Orbits}

\begin{lemma}[Error accumulation]
	\label{lem:accumulation_error}
    Let $x\in\RR^{d_x}$ with true orbit $\set{f^k(x)}_{k\ge0}$, and let $\set{y_k}_{k=0}^{n}$ be any sequence with $y_0=x$, with local errors $\delta_k\defeq\norm{y_{k+1}-f(y_k)}$ for $0\le k<n$. Write $E_k\defeq\norm{y_k-f^k(x)}$ for the iterated prediction error at lead $k$. If $f^k(x),y_k\in K$ for $0\le k<n$ (the final points $f^n(x),y_n$ need not lie in $K$), then
	\begin{equation}
		\label{eq:accumulation-error-exact}
		E_n \le \sum_{k=0}^{n-1}L^{n-1-k} \delta_k,
	\end{equation}
    and if $\delta_k \le \varepsilon$ for $0\le k<n$,
	\begin{equation}
		\label{eq:accumulation-error-bound}
		E_n\le B_L(n)\varepsilon,\qquad
		B_L(n) \defeq \sum_{j=0}^{n-1}L^j=\begin{cases} n, & L = 1, \\[4pt] \dfrac{L^n - 1}{L - 1}, & L \neq 1. \end{cases}
	\end{equation}
\end{lemma}

We call the right-hand side of \eqref{eq:accumulation-error-exact} the exact global bound, and $B_L(n)\varepsilon$ in \eqref{eq:accumulation-error-bound}, with $\varepsilon=\varepsilon_n$ the effective error of Definition \ref{def:local-error}, the effective global bound.

\begin{proof}
    This is the standard accumulation of local errors along a pseudo-orbit \citep{palmer2000shadowing}; we include the proof for completeness. For $0\le k<n$, since $y_k,f^k(x)\in K$, the triangle inequality and the Lipschitz condition give
	\begin{align*}
		E_{k+1} &= \norm{y_{k+1}-f\bigl(f^k(x)\bigr)} \\
		&\le \norm{y_{k+1}-f(y_k)} + \norm{f(y_k)-f\bigl(f^k(x)\bigr)}\\
		&\le\delta_k + L E_k.
	\end{align*}
	Applying the recursion repeatedly, at steps $k=n-1, n-2, \ldots, 0$ in turn, gives
    \begin{align*}
		E_n & \le \delta_{n-1}+L E_{n-1} \\
		&\le \delta_{n-1}+L\bigl(\delta_{n-2}+L E_{n-2}\bigr) = \delta_{n-1}+L\delta_{n-2}+L^{2}E_{n-2} \\
		&\; \; \vdots \\
		&\le \delta_{n-1}+L \delta_{n-2}+L^{2}\delta_{n-3}+\cdots+L^{ n-1}\delta_0+L^{n}E_0.
	\end{align*}
    Since $y_0=x$, we have $E_0=0$, and hence
	\[
		E_n \le \sum_{k=0}^{n-1}L^{n-1-k} \delta_k,
	\]
	which proves \eqref{eq:accumulation-error-exact}. If $\delta_k\le\varepsilon$ for $0\le k<n$, then, since $L^{n-1-k}\ge0$,
	\[
		E_n \le \varepsilon\sum_{k=0}^{n-1}L^{n-1-k}=\varepsilon\sum_{j=0}^{n-1}L^{j}=B_L(n) \varepsilon ,
	\]
	which is \eqref{eq:accumulation-error-bound}.
\end{proof}

Note that the proof uses nothing about $\set{y_k}_{k=0}^{n}$ beyond its local errors $\delta_k$ and its membership in $K$ for $k<n$, and nothing about the dynamics beyond the Lipschitz constant $L$ of $f$ on $K$. For the autonomous orbit of a pair $(x_t,r_t)$ the lemma is applied with $x=x_t$ and $y_k=\hat x_{t+k}$, so that $f^k(x)=x_{t+k}$ and $E_k=\norm{\hat x_{t+k}-x_{t+k}}$; the lead index $k$ of the lemma is the offset from the forecast origin. Two evaluations of the lemma are used repeatedly below. Evaluating it along the autonomous orbit of a true periodic point yields the residual bounds of Theorem \ref{thm:fixed_point_preservation}, and evaluating it along the autonomous orbit of an arbitrary pair $(x_t,r_t)$ yields the prediction-error bound of Theorem \ref{thm:error_propagation}.

\subsection{Periodic Points and Periodic Orbit Detection}
\label{sec:theory-fixedpoints}

\begin{theorem}[Periodic points of the autonomous reservoir]
	\label{thm:fixed_point_preservation}
	Let $n\ge1$ and $\varepsilon\ge0$.
	\begin{enumerate}\renewcommand{\labelenumi}{(\roman{enumi})}
		\item \textbf{True periodic points.} Let $x^*$ satisfy $f^n(x^*)=x^*$ (period dividing $n$), let $(x_t,r_t)$ be a pair with $x_t=x^*$, so that $r_t$ is a reservoir state that has just received $x^*$, and let $(\hat x_{t+k})$ be the autonomous orbit of $(x_t,r_t)$ with local errors $\delta_k$. Assume that the true and autonomous orbits remain in $K$ up to lead $n-1$, and that the local errors along them satisfy $\delta_k\le\varepsilon$ for $0\le k<n$. Then
		\begin{equation}
			\label{eq:thm-fp-bound}
			\norm{\fh{n}(x^*)-x^*}\le B_L(n) \varepsilon.
		\end{equation}
		In particular, if $x^*\in\Fix(f)$ then $\norm{\fh{1}(x^*)-x^*}=\delta_0\le\varepsilon$.

		\item \textbf{Periodic points of the reservoir.} Let $\tilde r\in\RR^{d_r}$ and let $\tilde y_k\defeq W_{\text{out}}G^{k}(\tilde r)$, $k\ge0$, be the outputs of the autonomous run from $\tilde r$. Suppose that the outputs $\set{\tilde y_k}$ and the true orbit of $\tilde y_0$ remain in $K$ up to lead $n-1$, and that the local errors $\norm{\tilde y_{k+1}-f(\tilde y_k)}$ of the outputs are each at most $\varepsilon$ for $0\le k<n$. Let $\kappa\defeq\norm{\tilde y_n-\tilde y_0} \le \norm{W_{\text{out}}}_2\norm{G^n(\tilde r)-\tilde r}$ be the closure gap. Then
		\begin{equation}
			\label{eq:thm-fp-residual-transfer}
			\norm{f^n(\tilde y_0)-\tilde y_0}\ \le\ B_L(n) \varepsilon+\kappa.
		\end{equation}

		\item \textbf{Exact reservoir cycles.} If in addition $\tilde r\in\Fix(G^n)$, then $\kappa=0$ and $\tilde y_0$ is a $B_L(n)\varepsilon$-approximate $n$-periodic point of $f$.
		\end{enumerate}
\end{theorem}

\begin{proof}
    (i) The sequence $y_k=\hat x_{t+k}$ satisfies the hypotheses of Lemma \ref{lem:accumulation_error} with $x=x^*$, and $f^n(x^*)=x^*$, so the lemma gives
	\[
		\norm{\fh{n}(x^*)-x^*}=\norm{\hat x_{t+n}-f^n(x^*)}\le B_L(n) \varepsilon,
	\]
	which is \eqref{eq:thm-fp-bound}. If $x^*\in\Fix(f)$, then $f(x^*)=x^*$, and the same identity at $n=1$ gives $\norm{\fh{1}(x^*)-x^*}=\norm{\hat x_{t+1}-f(x^*)}=\delta_0\le\varepsilon$.

    (ii) The outputs $\set{\tilde y_k}_{k=0}^{n}$ satisfy the hypotheses of Lemma \ref{lem:accumulation_error} with $x=\tilde y_0$ and $y_k=\tilde y_k$: the points $f^k(\tilde y_0)$ and $\tilde y_k$ lie in $K$ for $0\le k<n$, and the local errors are at most $\varepsilon$ by hypothesis. The lemma gives $\norm{\tilde y_n-f^n(\tilde y_0)}\le B_L(n)\varepsilon$, and the triangle inequality yields
	\[
		\norm{f^n(\tilde y_0)-\tilde y_0}\le\norm{f^n(\tilde y_0)-\tilde y_n}+\norm{\tilde y_n-\tilde y_0}\le B_L(n) \varepsilon+\kappa,
	\]
	which is \eqref{eq:thm-fp-residual-transfer}.

	(iii) If $\tilde r\in\Fix(G^n)$, then $\tilde y_n=W_{\text{out}}G^{n}(\tilde r)=W_{\text{out}}\tilde r=\tilde y_0$, so $\kappa=0$, and \eqref{eq:thm-fp-residual-transfer} reduces to $\norm{f^n(\tilde y_0)-\tilde y_0}\le B_L(n)\varepsilon$. By Definition \ref{def:approx-fixed-point}, $\tilde y_0$ is a $B_L(n)\varepsilon$-approximate $n$-periodic point of $f$.
\end{proof}

\begin{remark}[Residual versus distance to a periodic point]
    Part (ii) provides a bound on the residual $\norm{f^n(\tilde y_0)-\tilde y_0}$, but does not by itself provide a bound on the distance from $\tilde y_0$ to $\Fix(f^n)$: in general, a small residual does not imply that $\operatorname{dist}(\tilde y_0,\Fix(f^n))$ is small, and such an implication requires an additional error-bound condition. If in a neighborhood of the relevant periodic points there is $C>0$ with
	\begin{equation}
		\label{eq:local-error-bound-hyp}
		\operatorname{dist}(x,\Fix(f^n))\le C\norm{f^n(x)-x},
	\end{equation}
    then part (ii) gives, for $\tilde y_0$ in that neighborhood,
	\[
		\operatorname{dist}(\tilde y_0,\Fix(f^n))\le C \big(B_L(n)\varepsilon+\kappa\big).
	\]
	Condition \eqref{eq:local-error-bound-hyp} is an additional assumption, not implied by the standing assumptions.
\end{remark}

\begin{remark}[Direction of the Statement]
	Parts (ii)--(iii) of Theorem \ref{thm:fixed_point_preservation} go from $G$ to $f$: under the hypotheses of part (ii), periodic orbits of $G$ are approximate periodic orbits of $f$. The converse, that $G$ has a periodic orbit near every true one, would require additional assumptions on the model and is not established here.
\end{remark}

\begin{corollary}[Locating Approximate Periodic Points via the Autonomous Reservoir]
    \label{cor:periodic_detection}
    To locate approximate $n$-periodic points of $f$ using an ESN trained on $f$:
	\begin{enumerate}
		\item iterate the autonomous reservoir map $G$ $n$ times to obtain $G^n$;
		\item search for an approximate fixed point $\tilde r$ of $G^n$, satisfying $\norm{G^n(\tilde r)-\tilde r}\le\eta$, and set $\tilde y_0\defeq W_{\text{out}}\tilde r$;
		\item if $\tilde y_k\defeq W_{\text{out}}G^k(\tilde r)$ and $f^k(\tilde y_0)$ lie in $K$ and $\norm{\tilde y_{k+1}-f(\tilde y_k)}\le\varepsilon$ for $0\le k<n$, then, by Theorem \ref{thm:fixed_point_preservation}(ii), the residual for the true map satisfies $\norm{f^n(\tilde y_0)-\tilde y_0}\le B_L(n) \varepsilon+\norm{W_{\text{out}}}_2 \eta$.
	\end{enumerate}
	Steps 1 and 2 use only the trained reservoir; step 3 requires evaluating $f$ (and the constants $L$ and $\varepsilon$) and therefore cannot be carried out from data alone. Thus, the quality of the located periodic point is controlled by both the one-step approximation error $\varepsilon$ and the fixed-point residual $\eta$; for an exact fixed point of $G^n$ ($\eta=0$) this is Theorem \ref{thm:fixed_point_preservation}(iii).
\end{corollary}

\begin{remark}[Which Regime Governs the Residual Bound]
	The residual bound of Corollary \ref{cor:periodic_detection} inherits the two branches of $B_L(n)$: additive when $L\le1$, geometric when $L>1$. In the geometric branch the bound grows exponentially in $n$, so detection is most informative for small $n$.
\end{remark}

\subsection{Error Propagation and the Practical Prediction Horizon}
\label{sec:error-prop}

\begin{theorem}[Error Propagation Bounds for $\fh{n}$]
	\label{thm:error_propagation}
    Let $(x_t,r_t)$ be a pair and $n\ge1$. Assume that the true and autonomous orbits remain in $K$ up to lead $n-1$ (that is, $x_{t+k},\hat x_{t+k}\in K$ for $0\le k<n$), and that the local errors along them satisfy $\delta_k\le\varepsilon$ for $0\le k<n$. Then the iterated prediction error $E_n=\norm{\hat x_{t+n}-x_{t+n}}$ satisfies $E_n\le B_L(n)\varepsilon$, and in the three cases for $L$:
    \begin{enumerate}\renewcommand{\labelenumi}{(\roman{enumi})}
		\item \textbf{Contractive case} ($0\le L<1$): $E_n\le\varepsilon (1-L^n)/(1-L)\le\varepsilon/(1-L)$ for every $n$;
		\item \textbf{Neutral case} ($L=1$): $E_n\le n\varepsilon$;
        \item \textbf{Expansive case} ($L>1$): $E_n\le\varepsilon (L^n-1)/(L-1)$, a bound which grows geometrically in $n$.
    \end{enumerate}
\end{theorem}

For a set $\mathcal P$ of pairs, taking $\varepsilon=\varepsilon_n(\mathcal P)$ gives $B_L(n)\varepsilon_n(\mathcal P)$, which we call the uniform bound; it holds simultaneously for all pairs in $\mathcal P$.

\begin{proof}
    The bound $E_n\le B_L(n)\varepsilon$ is Lemma \ref{lem:accumulation_error} applied with $x=x_t$ and $y_k=\hat x_{t+k}$, so that $f^k(x)=x_{t+k}$, with $B_L(n)=\sum_{j=0}^{n-1}L^j$. (i) For $0\le L<1$ this is the geometric sum $(1-L^n)/(1-L)\le1/(1-L)$. (ii) For $L=1$ it equals $n$. (iii) For $L>1$ it equals $(L^n-1)/(L-1)$, which is at least its last term $L^{n-1}$ and so grows geometrically in $n$.
\end{proof}

\begin{remark}[Informative window]
	\label{rem:informative}
	If $x_{t+n},\hat x_{t+n}\in K$, then $E_n\le\operatorname{diam}(K)$ holds trivially, because no two points of $K$ are farther apart than $\operatorname{diam}(K)$. A bound with $B_L(n)\varepsilon\ge\operatorname{diam}(K)$ therefore tells us nothing new: it cannot exclude even the worst case, an error as large as $K$ itself. A bound is informative only if it lies below this worst case, which, for $L>1$, means $B_L(n)\varepsilon<\operatorname{diam}(K)$.
\end{remark}

We next relate the Lipschitz constant $L$ of the true map to the maximal Lyapunov exponent.

\begin{lemma}[Lipschitz Bound on the Maximal Lyapunov Exponent]
	\label{lem:lyapunov_lipschitz}
    Under the standing assumptions of Section \ref{sec:theory-setup}, suppose in addition that $f$ is $C^1$ on an open neighborhood of $K$ with $\norm{Df}_2\le L$ on $K$, and that the forward orbit of $x$ stays in $K$. Then
	\begin{equation}
		\lambda_{\max}(x)\defeq\limsup_{n\to\infty}\frac{1}{n} \ln\left(\norm{D(f^n)(x)}_2\right) \le \ln(L),
	\end{equation}
    or equivalently, $e^{\lambda_{\max}(x)} \le L$.
\end{lemma}

\begin{proof}
    By the chain rule, and since the forward orbit of $x$ stays in $K$ so that each $Df(f^k(x))$ is evaluated at a point of $K$ where $\norm{Df}_2\le L$,
	\[
		\norm{D(f^n)(x)}_2 \le \prod_{k=0}^{n-1}\norm{Df(f^k(x))}_2 \le L^n,
	\]
    so $(1/n)\ln\norm{D(f^n)(x)}_2 \le \ln(L)$ for every $n$. Therefore
	\[
		\lambda_{\max}(x)=\limsup_{n\to\infty}\frac{1}{n}\ln\norm{D(f^n)(x)}_2 \le \ln(L),
	\]
	and hence $e^{\lambda_{\max}(x)}\le L$.
\end{proof}

\noindent\textit{Lyapunov-rate estimate.}
The global bound uses the worst-case expansion $L$ over $K$, which need not reflect the expansion experienced along a particular orbit. The maximal Lyapunov exponent measures the asymptotic average exponential growth rate of perturbations along an orbit. Thus, if the one-step expansion along the orbit is represented, on average, by a factor $e^\lambda$ with $\lambda>0$, replacing $L$ by $e^\lambda$ in case (iii) of Theorem \ref{thm:error_propagation} suggests the practical estimate
\begin{equation}
	\label{eq:lyapunov-informal-bound}
	E_n \approx \varepsilon\frac{e^{n\lambda}-1}{e^\lambda-1} \approx \frac{\varepsilon}{e^\lambda-1}e^{n\lambda}, \qquad \text{for } n\lambda\gg1.
\end{equation}
The second approximation follows from the term $-1$ in the numerator becoming negligible once the exponential growth is sufficiently large. This estimate is not a consequence of Theorem \ref{thm:error_propagation} and is not a rigorous upper bound on $E_n$, since it replaces the step-by-step expansion by a single asymptotic average, whereas the actual expansion may fluctuate along the orbit. It is, however, no larger than the global bound based on $L$, since $e^\lambda\le L$ by Lemma \ref{lem:lyapunov_lipschitz}. A rigorous bound that accounts for the local expansion realized at every step is given by the following theorem.

\begin{theorem}[Local Jacobian Error Bound]
	\label{thm:local_jacobian}
	Assume in addition that $f$ is $C^1$ on an open neighborhood of $K$. Let $(x_t,r_t)$ be a pair and $n\ge1$, with true orbit $x_{t+k}=f^k(x_t)$, autonomous orbit $\set{\hat x_{t+k}}$ and iterated prediction errors $E_k=\norm{\hat x_{t+k}-x_{t+k}}$. Assume that the true and autonomous orbits remain in $K$ up to lead $n-1$, and that $\delta_k\le\varepsilon$ for $0\le k<n$. For $u,v\in K$ let $[u,v]\defeq\{(1-s)u+sv : 0\le s\le1\}\subset K$ be the connecting segment, and for $0\le k<n$ define the local Lipschitz bound on the segment joining the true and predicted state at lead $k$,
	\begin{equation}
		\ell_k \defeq \sup_{z\in[x_{t+k}, \hat x_{t+k}]} \norm{Df(z)}_2.
	\end{equation}
	Then
	\begin{equation}
		\label{eq:local-jacobian-bound}
		E_n \le \sum_{k=0}^{n-1}\delta_k\prod_{j=k+1}^{n-1} \ell_j\ \le\ \varepsilon \sum_{k=0}^{n-1} \prod_{j=k+1}^{n-1} \ell_j,
	\end{equation}
	with the empty product (at $k=n-1$) equal to $1$; in one dimension $\ell_j=\sup_{z\in[x_{t+j},\hat x_{t+j}]}\abs{f'(z)}$.
\end{theorem}

We call the first and second expressions in \eqref{eq:local-jacobian-bound}, the latter with $\varepsilon=\varepsilon_n$, the exact local bound and the effective local bound.

\begin{proof}
	We use the mean value inequality for continuously differentiable vector-valued maps \citep{rudin1976}: for $u,v\in K$,
	\begin{equation}
		\label{eq:mvi}
		\norm{f(u)-f(v)} \le \Big(\sup_{z\in[u,v]}\norm{Df(z)}_2\Big)\norm{u-v}.
	\end{equation}
	We now refine the recursion used to prove Lemma \ref{lem:accumulation_error}: instead of bounding $\norm{f(\hat x_{t+k})-f(x_{t+k})}$ by the global constant $L$, apply \eqref{eq:mvi} with $u=\hat x_{t+k}$, $v=x_{t+k}$, whose connecting segment has supremum $\ell_k$ by definition. Combined with the triangle inequality and the local error, exactly as in the proof of Lemma \ref{lem:accumulation_error}, this gives the refined recursion
	\begin{equation}
		\label{eq:rec}
		E_{k+1}\le\norm{\hat x_{t+k+1}-f(\hat x_{t+k})}+\norm{f(\hat x_{t+k})-f(x_{t+k})}\le\delta_k+\ell_kE_k,\qquad E_0=0.
	\end{equation}
    Applying \eqref{eq:rec} repeatedly, at steps $k=n-1, n-2, \ldots, 0$ in turn, gives
    \begin{align*}
        E_n &\le \delta_{n-1}+\ell_{n-1}E_{n-1} \\
		&\le \delta_{n-1}+\ell_{n-1}\bigl(\delta_{n-2}+\ell_{n-2}E_{n-2}\bigr) = \delta_{n-1}+\ell_{n-1}\delta_{n-2}+\ell_{n-1}\ell_{n-2}E_{n-2} \\
        &\; \; \vdots \\
        &\le\ \delta_{n-1}+\ell_{n-1}\delta_{n-2}+\ell_{n-1}\ell_{n-2}\delta_{n-3}+\cdots \\ &\quad +\bigl(\ell_{n-1}\ell_{n-2}\cdots\ell_{1}\bigr)\delta_0 +\bigl(\ell_{n-1}\ell_{n-2}\cdots\ell_{0}\bigr)E_0.
    \end{align*}
    Since $E_0=0$, the last term vanishes. The general term of the remaining sum is $\delta_k$ times the product of the local constants from lead $n-1$ down to lead $k+1$, that is, $\delta_k\prod_{j=k+1}^{n-1}\ell_j$ (an empty product when $k=n-1$), so
	\[
		E_n\ \le\ \sum_{k=0}^{n-1}\delta_k\prod_{j=k+1}^{n-1}\ell_j ,
	\]
	which is the first inequality in \eqref{eq:local-jacobian-bound}; the second uses $\delta_k\le\varepsilon$.
\end{proof}

\begin{remark}[Comparison with the global bound]
	Since $K$ is convex, the segment $[x_{t+k},\hat x_{t+k}]$ lies in $K$, so $\ell_k\le L$ for every $k$. Hence the first sum in \eqref{eq:local-jacobian-bound} is at most $\sum_{k=0}^{n-1}L^{n-1-k}\delta_k$, the bound \eqref{eq:accumulation-error-exact} of Lemma \ref{lem:accumulation_error}: the local bound is never larger than the global one, and equals it when $\ell_{k} = L$ for all $k$.
\end{remark}

\begin{remark}[Local bound and Lyapunov growth]
	When the predicted orbit remains close to the true orbit, $\ell_k$ reflects the local expansion along the true trajectory, so the products in \eqref{eq:local-jacobian-bound} describe the accumulated local expansion along it. This motivates comparison with finite-time Lyapunov growth, which is not a consequence of the theorem. The local bound can be substantially tighter than the global one when the trajectory rarely visits regions of maximal expansion.
\end{remark}

\begin{definition}[Practical prediction horizon]
	\label{def:prediction_horizon}
	Let $\lambda=\lambda_{\max}>0$ be the maximal Lyapunov exponent of $f$, and let $\varepsilon>0$ denote the effective error. The practical prediction horizon $n_{\max}$ is the lead at which the Lyapunov-rate estimate \eqref{eq:lyapunov-informal-bound} reaches $\operatorname{diam}(K)$, obtained by solving
	\[
		\varepsilon\frac{e^{n\lambda}-1}{e^\lambda-1} =\operatorname{diam}(K).
	\]
	Thus,
	\begin{equation}
		n_{\max} = \frac{1}{\lambda} \ln\left(1+\frac{(e^\lambda-1)\operatorname{diam}(K)}{\varepsilon}\right).
	\end{equation}
	When $\varepsilon \ll (e^\lambda-1)\operatorname{diam}(K)$, this gives
	\begin{equation}
		n_{\max} \approx \frac{1}{\lambda} \ln\left(\frac{\operatorname{diam}(K)}{\varepsilon}\right) + \frac{1}{\lambda} \ln(e^\lambda-1).
	\end{equation}
	The first term captures the leading dependence on the ratio $\operatorname{diam}(K)/\varepsilon$, scaled by $1/\lambda$. The second term accounts for the denominator $e^\lambda-1$ in the Lyapunov rate estimate. When this correction is negligible relative to the first term, the simpler approximation $n_{\max}\approx n_{\text{naive}}$, with
	\[
		n_{\text{naive}}\defeq\frac{1}{\lambda} \ln\left(\frac{\operatorname{diam}(K)}{\varepsilon}\right),
	\]
	is appropriate.
\end{definition}

The practical prediction horizon estimates when iterated prediction along a true orbit becomes unreliable, but is not a rigorous cutoff. Theorems \ref{thm:error_propagation} and \ref{thm:local_jacobian} remain valid beyond $n_{\max}$ whenever their hypotheses hold, and the actual error need not reach $\operatorname{diam}(K)$ at this lead. The exact expression shows that $n_{\max}$ increases as $\varepsilon$ decreases and decreases as $\lambda$ increases when $\varepsilon<\operatorname{diam}(K)$.

\section{Numerical Validation}
\label{sec:numerical}
This section evaluates the theory numerically on three systems of increasing dimension and difficulty. Section \ref{sec:val-protocol} describes the systems, reservoirs and evaluation criteria; Section \ref{sec:val-exp1} examines iterated prediction error, the global and local bounds, and the practical horizon; Section \ref{sec:val-exp2} examines residuals at true periodic points and fixed points; and Section \ref{sec:val-exp3} examines periodic orbits of the autonomous reservoir. Protocol details, exact reservoir hyperparameters and additional tables, including a summary of all checks, are in the Appendix.

\subsection{Systems, protocol, and standing diagnostics}
\label{sec:val-protocol}
The three systems differ in state dimension and attractor geometry, ranging from a one-dimensional unimodal map to a three-dimensional continuous-time flow. The first is the logistic map $f(x)=rx(1-x)$ with $r=3.91$ \citep{may1976}. The second is the H\'enon map $f(x,y)=(1-ax^2+y, bx)$ with $a=1.4$, $b=0.3$ \citep{henon1976}, a two-dimensional dissipative map with a folded attractor. The third is the Lorenz-63 system \citep{lorenz1963} with $\sigma_{\mathrm{L}}=10$, $\rho_{\mathrm{L}}=28$, $\beta_{\mathrm{L}}=8/3$, which we discretize with step size $h=0.09741577160869708$ using classical fourth-order Runge--Kutta with 20 substeps. The Jacobian $Df$ is computed exactly for the discretized map, using the same Runge--Kutta scheme. The step size is $h=T/16$, where $T=1.5586523$ is the period of the shortest periodic orbit of the numerical flow. Table \ref{tab:system-constants} shows that the one-step errors span six to seven orders of magnitude across the three systems, so a single set of experiments covers both very accurate and coarse one-step approximations.

For each system the domain $K$ is the observed bounding box of a 20{,}000-step true trajectory (after a 1{,}000-step transient period), inflated by 5\% of each coordinate's range, so $K$ is compact, convex and data-determined (see the Appendix). The Lipschitz constant is $L=\sup_{z\in K}\norm{Df(z)}_2$. For the logistic and H\'enon maps $Df$ is affine, so $z\mapsto\norm{Df(z)}_2$ is convex and its supremum over the box is attained at a vertex; $L$ and the segment constants $\ell_k$ of Theorem \ref{thm:local_jacobian} are then exact. For Lorenz both are evaluated on finite samples (box vertices and 20{,}000 uniform points for $L$; the endpoints and six interior points of each segment for $\ell_k$), so the Lorenz instances of the bounds are exact only up to this sampling. The maximal Lyapunov exponent $\lambda_{\max}$ is estimated by the QR/Benettin method \citep{benettin1980a,benettin1980b,wolf1985}; for Lorenz, the value in Table \ref{tab:system-constants} corresponds to $\lambda_{\max}/h=0.9066$ per unit time, in agreement with the commonly reported $\approx0.906$. The one-step error $\varepsilon_1$ is the worst case, that is the supremum over the forecast origins of the Euclidean one-step error $\delta_0$, in physical coordinates; it matches the uniform-bound hypothesis of Theorem \ref{thm:error_propagation} rather than a mean training error. The logistic fixed point $x=0$ and the Lorenz origin equilibrium are genuine fixed points of $f$ but lie outside $K$, so they are absent from the periodic-point searches and the corresponding reservoir roots are reported as non-admissible in Section \ref{sec:val-exp3}.

\begin{table*}
	\centering
	\caption{Measured worst-case one-step error $\varepsilon_1$, Lipschitz constant $L=\sup_{z\in K}\norm{Df(z)}_2$, maximal Lyapunov exponent $\lambda_{\max}$, domain diameter $\operatorname{diam}(K)$, and practical prediction horizon $n_{\max}$ of Definition \ref{def:prediction_horizon} with $\varepsilon=\varepsilon_1$. The quantities $\varepsilon_1$ and $n_{\max}$ refer to the representative seed.}
	\label{tab:system-constants}
	\begin{tabular}{@{}lrrrrrr@{}}
		\toprule
		System & $d_x$ & $\varepsilon_1$ & $L$ & $\lambda_{\max}$ & $\operatorname{diam}(K)$ & $n_{\max}$ \\
		\midrule
		Logistic ($r{=}3.91$) & 1 & $1.16\times10^{-8}$ & 4.083 & 0.4911 & 0.9807 & 36.2 \\
		H\'enon ($a{=}1.4,\,b{=}0.3$) & 2 & $6.67\times10^{-9}$ & 4.090 & 0.4233 & 2.937 & 45.5 \\
		Lorenz ($h{=}0.0974$) & 3 & $4.64\times10^{-2}$ & 4.900 & 0.0883 & 83.69 & 58.0 \\
		\bottomrule
	\end{tabular}
\end{table*}

Each reservoir is built and trained as in Section \ref{sec:training}; the reservoir dimension, spectral radius, sparsity, warm-up length and number of training pairs for each system are given in the Appendix. These are experimental choices, not values required by the theory: every theorem holds for any reservoir whose autonomous orbits satisfy its hypotheses. The readout is fitted by ordinary least squares without regularization, yielding very small one-step training root mean square errors of $1.6\times10^{-9}$, $3.4\times10^{-10}$ and $1.6\times10^{-4}$ for the logistic, H\'enon and Lorenz systems, respectively, in physical coordinates. However, the corresponding readout matrices have large spectral norms, $\norm{W_{\text{out}}}_2=2.2\times10^4$, $1.1\times10^4$ and $5.8\times10^3$, respectively, also in physical coordinates. These large norms are relevant to the root-finding tolerance used in Section \ref{sec:val-exp3}. The echo-state property holds by construction, since $W_{\text{res}}$ is symmetric with $\norm{W_{\text{res}}}_2=\rho<1$ for all 150 reservoirs, and is confirmed dynamically by the diagnostics of the Appendix.

Quantitative statements refer to one representative reservoir per system, selected from a sweep of 50 random seeds (0 to 49) per system as the one with the largest observed saturation lead (Section \ref{sec:val-exp1}), ties broken by smaller $\varepsilon_1$; the selected seeds are 15 for the logistic map, 37 for the Hénon map and 1 for the Lorenz system. The selection uses the same test origins that the figures display, so horizon-related figures are best-case, and the ratio of observed to predicted horizon at the selected seed is biased upwards; the sweep statistics are therefore reported alongside. The selection cannot bias any statement with theorem status, because the bound checks pass for every seed.

The forecast origins are the $n_{\text{lead}}=70$, $110$, $230$ steps of the data orbit immediately after the training segment, so they are out of sample; each origin is a pair $(x_t,r_t)$ as in Definition \ref{def:predictor}, with $r_t$ obtained by driving the reservoir with the true past from $r_0=0$. From every origin the reservoir runs autonomously for $n_{\text{lead}}$ leads. The pass/fail criteria are fixed before any result is shown. A case (a forecast origin at a lead, a periodic point, a point of a reservoir periodic orbit) is admissible if the hypotheses of the relevant theorem hold, that is, if the true and autonomous orbits remain in $K$ up to the required lead; only admissible cases are graded. A proven bound is violated if the measured quantity exceeds it by more than a relative tolerance of $10^{-9}$ plus an absolute tolerance of $10^{-12}\operatorname{diam}(K)$. Since the error cannot exceed $\operatorname{diam}(K)$, a bound above it carries no information; a bound below it is called non-vacuous (Remark \ref{rem:informative}), and we report non-vacuity alongside violation counts.

The Lyapunov-rate estimate \eqref{eq:lyapunov-informal-bound} is, as stated in Section \ref{sec:error-prop}, an indicative line and carries no pass/fail status.

\subsection{Error Propagation and the Practical Prediction Horizon}
\label{sec:val-exp1}
For every (origin, lead) pair we compare the iterated prediction error $E_n=\norm{\hat x_{t+n}-x_{t+n}}$ with five bounds. Two are global: the exact bound \eqref{eq:accumulation-error-exact} and the effective bound \eqref{eq:accumulation-error-bound}. Two are local: the exact and effective bounds, given by the first and second expressions of \eqref{eq:local-jacobian-bound}. The fifth is the uniform bound $B_L(n)\varepsilon_n(\mathcal P)$ of Theorem \ref{thm:error_propagation}, where $\varepsilon_n(\mathcal P)$ is the supremum of $\varepsilon_n$ over the admissible origins at lead $n$. For the representative seeds every origin and its autonomous orbit remain in $K$ at every lead tested, so all 69{,}900 cases are admissible (see the Appendix), and no bound is violated; nor is the ordering predicted after Theorem \ref{thm:local_jacobian}, that each local bound never exceeds the corresponding global bound. Repeating the experiment for all 50 seeds per system gives no violation and no ordering violation in any of the 150 reservoirs (3{,}486{,}549 admissible (origin, lead) cases, which are strongly dependent; all cases are admissible for the two maps and 99.7\% for Lorenz; see the Appendix). The one-step error ranges over almost nine orders of magnitude across systems and seeds ($5.5\times10^{-10}$ to $2.1\times10^{-1}$, see the Appendix) and the bounds hold throughout.

To confirm the checks can fail, we repeated the bound checks with $L$ and each $\ell_k$ scaled by a factor chosen to remain below the realized expansion (0.3 for the maps, 0.1 for Lorenz). This produced violations in 91.1\% of 139{,}800 checks (see the Appendix). The true constants therefore pass checks that constants contradicted by the data fail.

Validity and tightness are different questions, and the central quantitative finding of this experiment concerns the regime in which each proven bound is informative. Table \ref{tab:window} lists the last lead at which each bound stays below $\operatorname{diam}(K)$. The local bound extends the informative window of the median global bound by factors of 2.3, 1.8 and 3.3 for the logistic, H\'enon and Lorenz systems, and that of the uniform bound, which is governed by the worst origin, by factors of 2.5, 2.1 and 5.2 (ratios of the entries of Table \ref{tab:window}). This is consistent with Theorem \ref{thm:local_jacobian}: the local bound tracks the derivatives realized along the orbit instead of the global worst case, and all three orbits spend most of their time far from regions of maximal expansion. The improvement is largest for Lorenz, where $L$ far exceeds $e^{\lambda_{\max}}$ (Table \ref{tab:system-constants}). The local bound is nevertheless conservative: at $n=10$ it exceeds the median error by factors of about 16, 94 and 34, while the median global bound exceeds the median local bound by about $8\times10^2$, $3\times10^2$ and $1.4\times10^4$ (see the Appendix). Figure \ref{fig:bounds} shows the same contrast origin by origin, most visibly for Lorenz, where the global bound leaves the plotted range within a dozen leads.

\begin{table*}
	\centering
	\caption{Informative window of the proven bounds (representative seeds). Left: last lead at which a bound is below $\operatorname{diam}(K)$, for the median (over origins) effective global bound, the uniform bound $B_L(n)\varepsilon_n(\mathcal P)$ (set by the worst origin), and the median effective local bound. Right: fraction of all (origin, lead) cases in which the effective global or local bound is below $\operatorname{diam}(K)$.}
	\label{tab:window}
	{\small\setlength{\tabcolsep}{2.5pt}
		\begin{tabular}{@{}lrrrrr@{}}
			\toprule
			& \multicolumn{3}{c}{Last lead below $\operatorname{diam}(K)$} & \multicolumn{2}{c}{Cases below $\operatorname{diam}(K)$} \\
			\cmidrule(lr){2-4}\cmidrule(l){5-6}
			System & global (median) & global (uniform) & local (median) & global & local \\
			\midrule
			Logistic & 12 & 11 & 27 & 17\% & 40\% \\
			H\'enon & 16 & 14 & 29 & 14\% & 26\% \\
			Lorenz & 8 & 5 & 26 & 3\% & 11\% \\
			\bottomrule
		\end{tabular}
	}
\end{table*}

\begin{figure*}
    \centering
    \includegraphics[width=\textwidth]{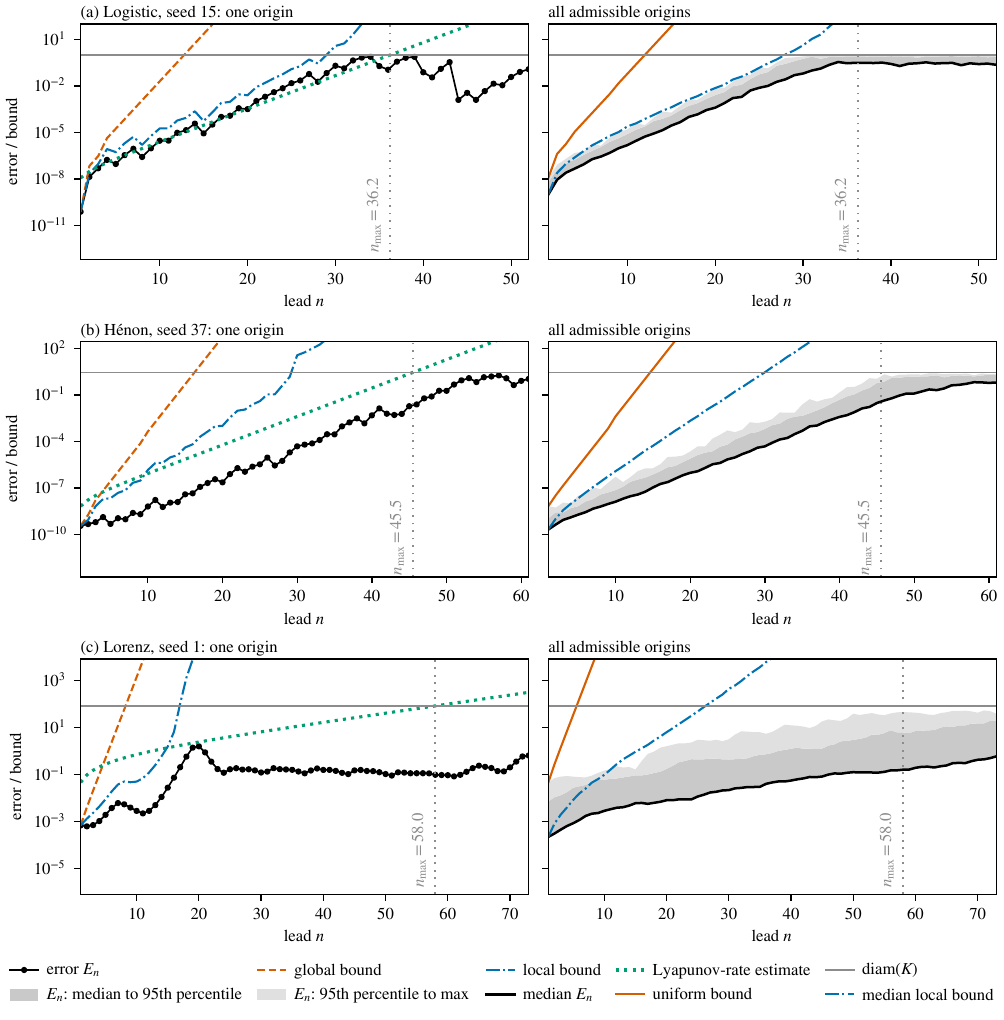}
    \caption{(a) Logistic map (seed 15), (b) H\'enon map (seed 37), (c) Lorenz system (seed 1). Iterated prediction error $E_n$ against the theoretical reference curves, leads $1$ to $\lceil n_{\max}\rceil+15$, logarithmic scale. Left panels: one representative origin (the origin with median observed saturation lead among those admissible at the final lead), showing the error, the effective global bound $B_L(n)\varepsilon_n$, the effective local bound, the Lyapunov-rate estimate with $\varepsilon=\varepsilon_1$, $\operatorname{diam}(K)$, and the horizon $n_{\max}$ (dotted vertical line, labeled with its value). Right panels: all admissible origins, showing the median error with the band from median to 95th percentile and from 95th percentile to maximum, the uniform bound $B_L(n)\varepsilon_n(\mathcal P)$ of Theorem \ref{thm:error_propagation}, and the median effective local bound.}
	\label{fig:bounds}
\end{figure*}

\begin{figure*}
    \centering
    \includegraphics[width=0.94\textwidth]{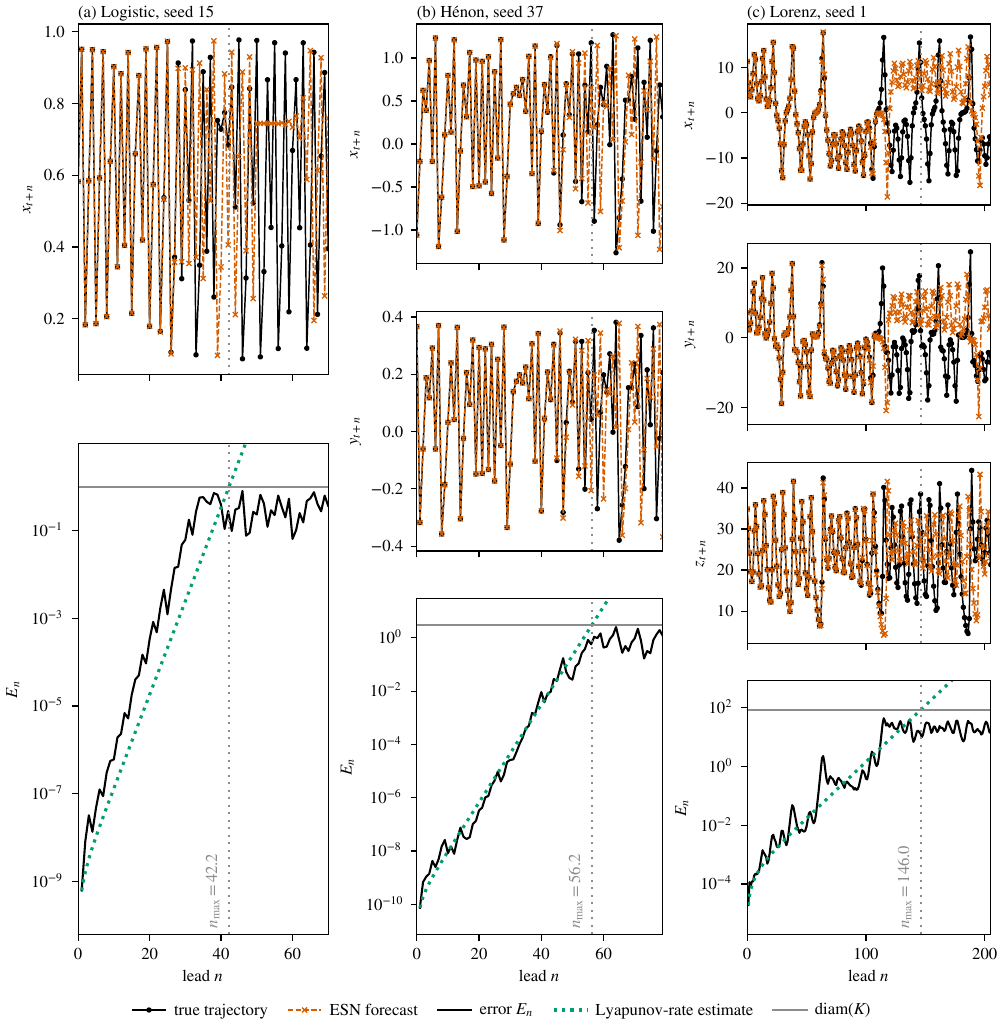}
    \caption{(a)--(c) Autonomous forecasts from typical origins, one column per system (logistic $r{=}3.91$; H\'enon $a{=}1.4$, $b{=}0.3$; Lorenz $h{=}0.0974$, state components $x$, $y$, $z$). Top rows: the true orbit (dots) and the autonomous ESN forecast (crosses) against lead $n$; the dotted vertical line, labeled with its value, is the horizon $n_{\max}$ of Definition \ref{def:prediction_horizon} computed from this origin's own one-step error, which is why it differs from the sup-based values of Table \ref{tab:system-constants}. Bottom row: the error $E_n$ (solid) on a logarithmic scale against the Lyapunov-rate estimate $\varepsilon(e^{n\lambda_{\max}}-1)/(e^{\lambda_{\max}}-1)$ built from the same one-step error (dotted), with $\operatorname{diam}(K)$ for scale. The origin shown is the one whose ratio of observed saturation lead to its own $n_{\max}$ is the median among the forecast origins. The reservoir seed of each system is given in the panel title.}
	\label{fig:forecast}
\end{figure*}

\noindent\textit{The practical horizon.} Definition \ref{def:prediction_horizon} is a heuristic and we test it as such. Figure \ref{fig:forecast} shows that for a typical origin the error follows the Lyapunov-rate estimate, built from that origin's own one-step error, for several orders of magnitude and then saturates at the scale of the attractor, after which forecast and true trajectory are uncorrelated. The horizon marks where iterated prediction against a specific orbit ceases to be informative, not where the proven bounds cease to hold.

Table \ref{tab:horizon} calibrates the horizon against the data. The observed saturation lead $n_{\mathrm{emp}}$ is the first lead at which the median error over origins reaches $10\%$ of $\operatorname{diam}(K)$. Since $n_{\max}$ targets $100\%$ of $\operatorname{diam}(K)$, we also report $n_{0.1}$, the lead at which the same estimate reaches $10\%$, and $n_{0.1}^{\mathrm{med}}$, which additionally replaces the worst-case $\varepsilon_1$ by the median one-step error over origins, since the observed median error is driven by the typical rather than the worst origin. The approximation $n_{\text{naive}}$ of Definition \ref{def:prediction_horizon}, evaluated at $\varepsilon=\varepsilon_1$, overestimates $n_{\max}$ by approximately $-\lambda_{\max}^{-1}\ln(e^{\lambda_{\max}}-1)$; this is negligible for the maps (about 1 and 1.5 leads, Table \ref{tab:horizon}) but reaches 27 leads, a 46\% overestimate, for Lorenz, whose small exponent makes the correction large, so the exact formula of the definition should be used.

For the two maps the estimate is accurate: $n_{\mathrm{emp}}/n_{0.1}$ is 1.01 and 1.37, and 0.88 and 1.14 with the median one-step error (Table \ref{tab:horizon}). For the Lorenz system the sup-based estimate is conservative by a factor of 3.5, largely because the worst-case one-step error is 207 times its median ($4.6\times10^{-2}$ against $2.2\times10^{-4}$); with the median error the ratio falls to 1.25, so all three calibrated ratios lie between 0.88 and 1.25. The horizon is thus a calibration of an indicative estimate and not a proven cutoff, and prediction remains informative past the sup-based $n_{\max}$, by a factor of about two for the Lorenz system ($n_{\mathrm{emp}}/n_{\max}=115/58$, Table \ref{tab:horizon}).

\begin{table*}
	\centering
	\caption{Horizon calibration (representative seeds). $n_{\max}$: horizon of Definition \ref{def:prediction_horizon} with $\varepsilon=\varepsilon_1$; $n_{\text{naive}}$: simpler approximation of Definition \ref{def:prediction_horizon} with $\varepsilon=\varepsilon_1$; $n_{0.1}$: lead at which the Lyapunov-rate estimate with $\varepsilon=\varepsilon_1$ reaches $10\%\operatorname{diam}(K)$; $n_{0.1}^{\mathrm{med}}$: the same with $\varepsilon$ the median one-step error over origins (logistic $1.0\times10^{-9}$, H\'enon $2.1\times10^{-10}$, Lorenz $2.2\times10^{-4}$); $n_{\mathrm{emp}}$: first lead at which the median error over origins reaches $10\%\operatorname{diam}(K)$. The last two columns are the ratios of $n_{\mathrm{emp}}$ to $n_{0.1}$ and $n_{0.1}^{\mathrm{med}}$.}
	\label{tab:horizon}
	\begin{tabular}{@{}lrrrrrrrr@{}}
		\toprule
		System & $\lambda_{\max}$ & $n_{\max}$ & $n_{\text{naive}}$ & $n_{0.1}$ & $n_{0.1}^{\mathrm{med}}$ & $n_{\mathrm{emp}}$ & $n_{\mathrm{emp}}/n_{0.1}$ & $n_{\mathrm{emp}}/n_{0.1}^{\mathrm{med}}$ \\
		\midrule
		Logistic ($r{=}3.91$) & 0.4911 & 36.2 & 37.2 & 31.5 & 36.6 & 32 & 1.01 & 0.88 \\
		H\'enon ($a{=}1.4,\,b{=}0.3$) & 0.4233 & 45.5 & 47.0 & 40.1 & 48.3 & 55 & 1.37 & 1.14 \\
		Lorenz ($h{=}0.0974$) & 0.0883 & 58.0 & 84.9 & 32.5 & 92.2 & 115 & 3.54 & 1.25 \\
		\bottomrule
	\end{tabular}
\end{table*}

\subsection{True Periodic Points and Fixed-Point Residuals}
\label{sec:val-exp2}
True periodic points of $f^n$ are located by numerical root finding (see the Appendix), and the roots found are filtered by membership in $K$ and deduplicated; the roots are accurate to $8.8\times10^{-13}$ in $\norm{f^n(x)-x}$, and this accuracy is added as slack to every bound checked at that point. The search finds 18 true points of $f^n$ for the logistic map across orders $n=1$ to 4, 16 for H\'enon, and for Lorenz the two equilibria at $n=1$ plus the orbit of prime period 16 described below. For the logistic and H\'enon maps, independent exact counts for $n\le4$ agree with the numbers found (see the Appendix): the logistic map has 2, 4, 8, 8 real fixed points of $f^n$, all of which except $x=0$ lie in $K$, giving 1, 3, 7, 7, and H\'enon has 2, 4, 2, 8, so there is no real point of prime period 3 at these parameters.

For Lorenz we also test order $n=16$: the step size $h=T/16$ was chosen so that the shortest periodic orbit of the numerical flow becomes an orbit of prime period 16 of $f$. The 16-fold composed map is then the time-$T$ flow map, so every point of this orbit satisfies $f^{16}(x)=x$ (to $8.8\times10^{-13}$), and $\Fix(f^{16})\cap K$ is the union of a closed curve and the two equilibria, rather than a finite set. The 16 test points used at this order are the 16 points of one orbit of prime period 16 of $f$ on this curve, evenly spread in phase. This is also why one-to-one matching of points is replaced, for this order, by a comparison of whole orbits in Section \ref{sec:val-exp3}.

For each true periodic point $x^*$ the reservoir is driven along the true periodic orbit (repeated to at least 1{,}000 steps and ending at $x^*$), so that its state is synchronized and has just received $x^*$, and is then run autonomously for $n$ steps. All 52 points are admissible and the residual $\norm{\fh{n}(x^*)-x^*}$ satisfies all five bounds, of which 212 of the 260 individual checks are non-vacuous (see the Appendix). Figure \ref{fig:periodic} shows each point of prime period $n$ against the tightest bound. The more general bounds are looser. For Lorenz at $n=16$ only the local bounds are informative: with $L=4.9$ (Table \ref{tab:system-constants}) the global factor is $B_L(16)=(L^{16}-1)/(L-1)\approx3\times10^{10}$, so the global bound exceeds $\operatorname{diam}(K)=83.7$ for any $\varepsilon\gtrsim3\times10^{-9}$, whereas the tightest bound in Figure \ref{fig:periodic} lies below $\operatorname{diam}(K)$ for all 16 points of the orbit. The content of the result at this order is carried by the local bound of Theorem \ref{thm:local_jacobian}, as Remark \ref{rem:informative} anticipates once $B_L(n)\varepsilon$ outgrows the diameter.

\begin{figure*}
    \centering
    \includegraphics[width=\textwidth]{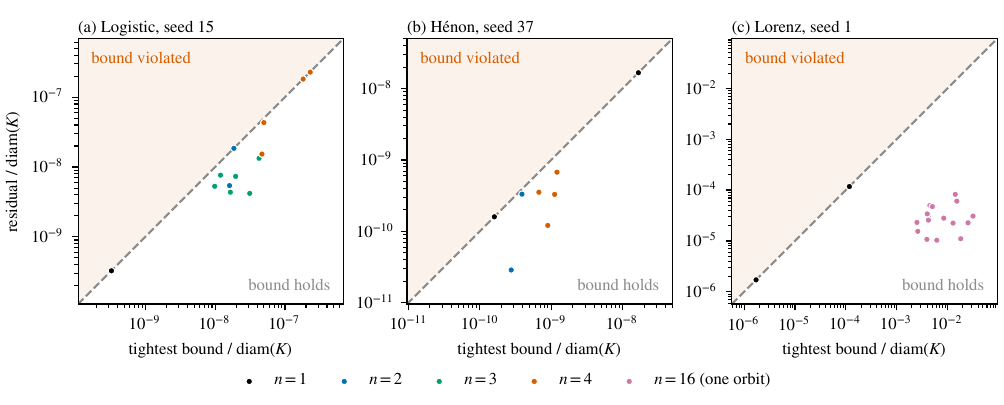}
    \caption{Theorem \ref{thm:fixed_point_preservation}(i) at every true periodic point of prime period $n$ (logistic: 13 points; H\'enon: 8; Lorenz: 18, of which 16 lie on the closed orbit of prime period 16). Each dot is one point, plotted at the tightest of the five bounds (horizontal) and the realized residual $\norm{\fh{n}(x^*)-x^*}$ (vertical), both normalized by $\operatorname{diam}(K)$, on log-log axes with the diagonal $y=x$. The theorem asserts that every dot lies on or below the diagonal (up to the numerical tolerance of Section \ref{sec:val-protocol}); the distance below it is the slack. The $n=1$ points (black) sit on the diagonal because at $n=1$ the bounds equal $\delta_0$ identically. The reservoir seed of each system is given in the panel title.}
	\label{fig:periodic}
\end{figure*}

Each genuine fixed point of $f$ (one for the logistic map, two each for H\'enon and Lorenz) is also used as a forecast origin, with the reservoir state obtained by driving with the constant input $x^*$ for 1{,}000 steps, and the residual $\norm{\hat x_{t+n}-x^*}$ is tracked against the pointwise bounds at every lead up to $n_{\text{lead}}$. All 750 (starting point, lead) cases are admissible and none violates any bound (see the Appendix). Figure \ref{fig:fpresidual} shows the growth curves, and three features carry the interpretation. For the logistic and H\'enon maps (first two panels of Figure \ref{fig:fpresidual}) the residual grows geometrically at about the rate $e^{\lambda_{\max}}$, follows the local bound at first (for the H\'enon point $x_1^*$ it nearly touches it, as expected since $\ell_k\approx\norm{Df}$ along the nearby true fixed orbit), and then saturates at a substantial fraction of the domain. The residuals are not monotone, since the autonomous orbit returns intermittently toward $x^*$; this is immaterial for validity because the bounds are worst-case envelopes. And over the plotted leads the Lorenz residuals do not grow exponentially but oscillate orders of magnitude below the attractor scale, with the local bound crossing $\operatorname{diam}(K)$ only at $n\approx38$ (dotted line in the Lorenz panel of Figure \ref{fig:fpresidual}). This follows from the linearization of the time-$h$ map at the equilibria, which are saddle-foci with one real contracting direction (multiplier $0.259$) and a complex pair of modulus $1.0092$ and argument $0.993$ rad: perturbations grow by only $0.9\%$ per step while rotating with period $2\pi/0.993\approx6.3$ steps, matching the oscillation in the Lorenz panel of Figure \ref{fig:fpresidual}. The tested leads reach 1.9, 2.4 and 4.0 times each system's $n_{\max}$ ($n_{\text{lead}}$ in the Appendix, $n_{\max}$ in Table \ref{tab:system-constants}) with every bound intact; the horizon concerns iterated prediction against a specific orbit, whereas fixed-point recovery is a zeroth-order question of self-consistency of $\fh{n}$ near $x^*$, which degrades far more gently when the true orbit is nearly stationary.

\begin{figure*}
    \centering
    \includegraphics[width=\textwidth]{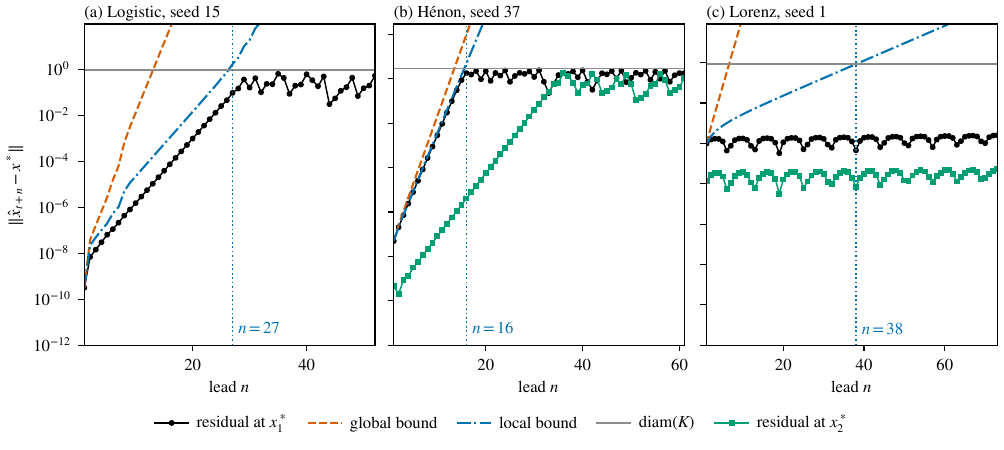}
    \caption{Residual $\norm{\hat x_{t+n}-x^*}$ from each true fixed point of $f$ under autonomous iteration (markers), against the effective global bound $B_L(n)\varepsilon_n$ (dashed) and the effective local bound (dash-dotted, shown for $x_1^*$), for leads 1 to $\lceil n_{\max}\rceil+15$, with $\operatorname{diam}(K)$ for scale. The dotted vertical line marks the first lead at which the local bound exceeds $\operatorname{diam}(K)$. Logistic: one fixed point; H\'enon: two; Lorenz: two (the flow equilibria). The reservoir seed of each system is given in the panel title.}
	\label{fig:fpresidual}
\end{figure*}

\subsection{Periodic Orbits of the Autonomous Reservoir and Their Detection}
\label{sec:val-exp3}
This subsection examines whether the autonomous reservoir recovers the periodic points of the target. Fixed points of the composed autonomous map $G^n$ are located by root finding in reservoir space alone and compared with the true periodic points of $f^n$. The roots found are approximate, with residual $\eta=\norm{G^n(r)-r}$, which the theory accommodates: $\eta$ enters the bound of Corollary \ref{cor:periodic_detection} through the term $\norm{W_{\text{out}}}_2\eta$. The bounds apply to every root whose outputs and true orbit stay in $K$ (Theorem \ref{thm:fixed_point_preservation}(ii)), however the root was obtained.

\begin{figure*}
    \centering
    \includegraphics[width=\textwidth]{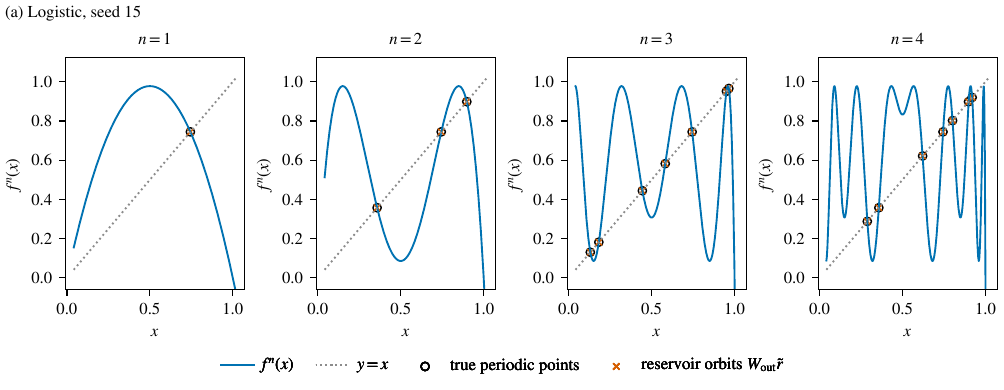}
    \\[1ex]
    \includegraphics[width=\textwidth]{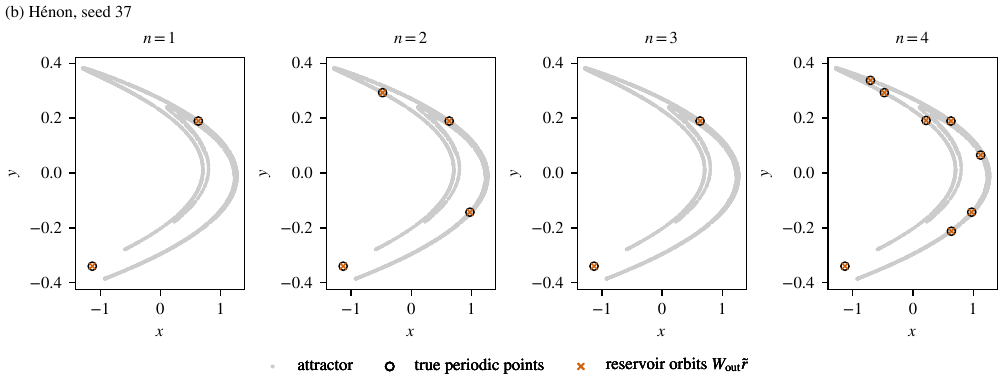}
    \caption{True periodic points of $f^n$ (open circles) and the outputs $W_{\text{out}}\tilde r$ of the admissible reservoir orbit points (crosses), $n=1$ to $4$. (a) Logistic map: the graphs of $f^{n}$ and the identity, with fixed points of $f^{n}$ on the diagonal. (b) H\'enon map: both periodic points and reservoir outputs in the $(x,y)$-plane, shown against the attractor. The reservoir seed of each system is given in the panel title.}
\label{fig:orbits}
\end{figure*}

The searches find 105 roots of $G^n$ across the three systems, of which 61 are admissible (Table \ref{tab:orbit-search}; the outputs $y_k=W_{\text{out}}G^k(\tilde r)$ and the true orbit of $y_0$ stay in $K$ through lead $n-1$). Every admissible root is matched one-to-one to a true periodic point of the same order (Table \ref{tab:orbit-search} and Figure \ref{fig:orbits}: 18 of 18 for the logistic map, 16 of 16 for H\'enon, 2 of 2 for Lorenz at $n=1$) and none is spurious. Since the true sets are complete for the two maps (Section \ref{sec:val-exp2}), ``the reservoir recovers every periodic point of order $n\le4$'' is a complete statement for them. It is an empirical observation, not a theorem: the converse direction, that $G$ has a periodic orbit near every true periodic point, is not established here. The proven inequalities also hold for all admissible roots: the transfer bound $\norm{f^n(y_0)-y_0}\le B_L(n)\varepsilon+\kappa$ of Theorem \ref{thm:fixed_point_preservation}(ii), the detection bound of Corollary \ref{cor:periodic_detection}, and the Lemma \ref{lem:accumulation_error} and Theorem \ref{thm:local_jacobian} bounds on $E_n=\norm{y_n-f^n(y_0)}$; the closure-gap inequality $\kappa\le\norm{W_{\text{out}}}_2\eta$ holds for all 105 roots. The Corollary bound is valid but much weaker than part (ii): the realized closure gap never exceeds 7.6\% of its allowance $\norm{W_{\text{out}}}_2\eta$ for admissible roots (for one logistic root of $G^2$ the part-(ii) bound is about $5\times10^{-8}$ against $5\times10^{-4}$ for the Corollary). This looseness is a consequence of the unregularized readout; a regularized $W_{\text{out}}$ would reduce the gap at the cost of a larger one-step error.

The remaining 44 roots are genuine periodic orbits of the composed ESN map whose output orbits leave $K$, so the theory makes no claim about them and they are not counted as violations. Most have outputs far outside $K$, typically with saturated reservoir components; the others are reservoir images of genuine fixed points of $f$ outside $K$, namely $x=0$ for the logistic map ($y_0\approx-9\times10^{-6}$) and, for Lorenz, a root at $(0,\,0.013,\,-0.122)$ near the origin equilibrium. Their presence suggests, without proof, that the reservoir has captured equilibria of $f$ outside the range of the training data. We do not study them further, for instance through their residuals from the true fixed points or from true periodic points, because the constants $L$, $\ell_k$ and $\varepsilon_n$ are defined on $K$ and the reservoir is trained only on data from $K$, so there is neither a bound to test nor a measured one-step error outside it.

\begin{table*}
	\centering
	\caption{Periodic-orbit search by system and order (representative seeds). Found: roots of $G^n(r)-r$ after deduplication; adm.: admissible roots; matched/true: admissible outputs matched one-to-one (Hungarian assignment, tolerance $5\%$ of $\operatorname{diam}(K)$) to the true periodic points of Section \ref{sec:val-exp2}; max $\eta$: largest root residual $\norm{G^n(r)-r}$, measured in the reservoir state space. No admissible output is spurious, meaning that every admissible output is matched to a root, and no bound is violated at any admissible root. For Lorenz $n=16$ the true points of prime period 16 form a closed curve, so matching is replaced by the orbit comparison of Table \ref{tab:lorenz-orbits}; its 25 admissible roots are the 23 orbit points of that table and the two equilibria.}
	\label{tab:orbit-search}
	{\small
		\begin{tabular}{@{}llrrrr@{}}
			\toprule
			System & $n$ & found & adm. & matched/true & max $\eta$ \\
			\midrule
			Logistic & 1 & 4 & 1 & 1/1 & $3.5\times10^{-8}$ \\
			& 2 & 6 & 3 & 3/3 & $2.7\times10^{-7}$ \\
			& 3 & 9 & 7 & 7/7 & $6.1\times10^{-7}$ \\
			& 4 & 9 & 7 & 7/7 & $8.3\times10^{-7}$ \\
			\addlinespace
			H\'enon & 1 & 3 & 2 & 2/2 & $2.5\times10^{-8}$ \\
			& 2 & 8 & 4 & 4/4 & $4.5\times10^{-8}$ \\
			& 3 & 9 & 2 & 2/2 & $6.9\times10^{-7}$ \\
			& 4 & 10 & 8 & 8/8 & $9.5\times10^{-8}$ \\
			\addlinespace
			Lorenz & 1 & 22 & 2 & 2/2 & $8.7\times10^{-7}$ \\
			& 16 & 25 & 25 & n/a & $4.6\times10^{-7}$ \\
			\bottomrule
		\end{tabular}
	}
\end{table*}

\begin{figure*}
    \centering
    \includegraphics[width=\textwidth]{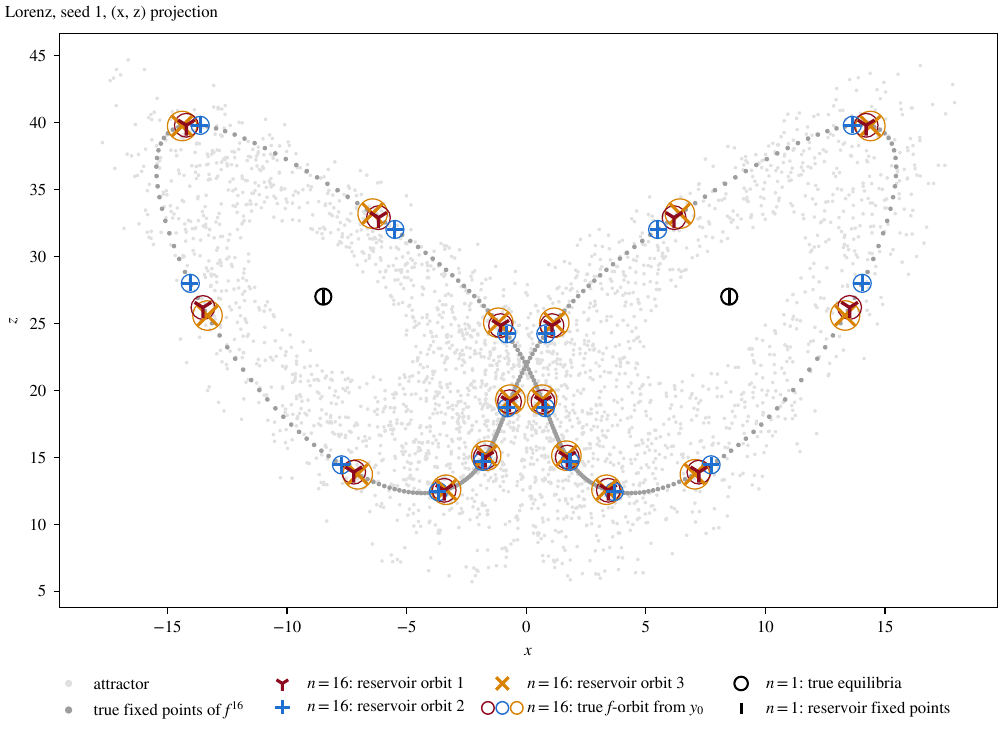}
    \caption{True periodic points of $f^n$ and reservoir orbits for the Lorenz system, shown in the $(x,z)$-projection. At $n=1$, the fixed points are the two equilibria (black). At $n=16$, $\Fix(f^{16})\cap K$ consists of the closed orbit of prime period 16 (small dots) together with the two equilibria; the reservoir orbits found are numbered as in Table \ref{tab:lorenz-orbits} and drawn with distinct markers and colors (orbit 1: dark red tri-down mark; orbit 2: blue $+$; orbit 3: orange $\times$), each against the true $f$-orbit started at the same $y_0$ (open circles in the same color). At $n=1$ the open black circles are the true equilibria and the black vertical bars are the reservoir fixed points. The reservoir seed is given in the figure title.}
    \label{fig:orbits-lorenz}
\end{figure*}

The Lorenz case $n=16$, where the periodic points form a closed curve, is the most distinctive result. The transfer bounds at $n=16$ are valid but vacuous ($B_L(16)\approx3\times10^{10}$ for $L=4.9$, Table \ref{tab:system-constants}, against $\operatorname{diam}(K)=83.7$; see Section \ref{sec:val-exp2}), so the evidence is geometric: how closely the recovered orbits follow the true closed orbit. The 25 admissible roots at this order decompose into three orbits of prime period 16 under $G$ (Table \ref{tab:lorenz-orbits}, Figure \ref{fig:orbits-lorenz}) together with the two equilibria already matched at $n=1$, and no root is left over. Every output of each orbit lies within $6.7\times10^{-4}$ to $8.0\times10^{-4}$ of $\Fix(f^{16})$ (column ``max dist.'' of Table \ref{tab:lorenz-orbits}; at most $9.5\times10^{-6}\operatorname{diam}(K)$), and each orbit follows the true $f$-orbit started at its own $y_0$ to within $1.3\times10^{-3}$ to $3.4\times10^{-3}$ (Table \ref{tab:lorenz-orbits}). The stability multipliers of $DG^{16}$ along each orbit (last column of Table \ref{tab:lorenz-orbits}) recover those of the true orbit: the leading multipliers are $4.708$, $4.713$ and $4.708$ against the true $4.713$ (errors at most $0.10\%$), and the middle multipliers lie within $0.11\%$ of $1$, the structurally correct result since the tangent direction along an orbit that shadows a periodic orbit of a flow-sampled map is neutral. The contracting multipliers ($0.018$ against the true $1.2\times10^{-10}$) are the one first-order quantity the reservoir does not resolve: it attenuates transverse contraction at a per-step factor of about $e^{-0.25}$ where the true map contracts at $e^{-1.43}$ (the sixteenth roots of the contracting multipliers in Table \ref{tab:lorenz-orbits}). Newton coverage of the orbits varies (12, 10 and 1 of the 16 points found as separate roots; column ``Roots'' of Table \ref{tab:lorenz-orbits}), but each orbit is verified as a whole by orbit-following, since one point determines the whole orbit. For this seed, a reservoir trained on 300 one-step Lorenz pairs ($T_1-T_0$ in the Appendix) has orbits of prime period 16 within $8\times10^{-4}$ of the shortest periodic orbit of the discretized flow, found by root finding on $G^{16}$.

\begin{table*}
\centering
\caption{Lorenz, $n=16$ (seed 1), where $\Fix(f^{16})\cap K$ contains a closed curve: the three reservoir orbits found against the true closed orbit $\Fix(f^{16})$. Roots are points of the orbit found as separate Newton roots. The maximum distance of the orbit outputs from the closed curve is reported, along with the tracking error, defined as the largest deviation of the orbit outputs from the true $f$-orbit started at the orbit's own $y_0$. The multipliers are the absolute eigenvalues of $DG^{16}$ along the orbit, compared with the true orbit's values $4.7129$, $1.0000$, and $1.2\times10^{-10}$. Distances are in physical units, with $\operatorname{diam}(K)=83.7$.}
	\label{tab:lorenz-orbits}
	\begin{tabular}{@{}crrrrl@{}}
    \toprule
    \makecell{Orbit} & \makecell{Prime period\\under $G$} & \makecell{Roots} & \makecell{max dist.} & \makecell{tracking} & \makecell{multipliers of $DG^{16}$} \\
		\midrule
		1 & 16 & 12/16 & $6.9\times10^{-4}$ & $2.3\times10^{-3}$ & $4.7080, 1.0007, 0.0179$ \\
		2 & 16 & 10/16 & $6.7\times10^{-4}$ & $3.4\times10^{-3}$ & $4.7126, 0.9993, 0.0184$ \\
		3 & 16 & 1/16 & $8.0\times10^{-4}$ & $1.3\times10^{-3}$ & $4.7081, 1.0011, 0.0176$ \\
		\bottomrule
	\end{tabular}
\end{table*}

\section{Discussion and Conclusions}
\label{sec:discussion}
We have shown that the autonomous $n$-step predictors $\fh{n}$ of a reservoir trained solely on the one-step map $f$ carry provable information about $f^n$. Under explicit hypotheses, the $n$-step prediction started at a true point of period $n$ returns close to it, and approximate periodic points of the autonomous reservoir map are approximate periodic points of $f$, which yields a residual-based procedure for locating approximate periodic points (Theorem \ref{thm:fixed_point_preservation}, Corollary \ref{cor:periodic_detection}); the prediction error obeys a global geometric bound and a sharper local Jacobian bound (Theorems \ref{thm:error_propagation} and \ref{thm:local_jacobian}); and a Lyapunov-rate heuristic defines a practical horizon $n_{\max}$ (Definition \ref{def:prediction_horizon}). No proven bound was violated in 70{,}763 admissible checks for the representative reservoirs or in 3{,}486{,}549 checks across 50 realizations per system (see the Appendix), whereas deliberately reduced constants failed in 91.1\% of checks (see the Appendix). The local bound remained informative for 1.8 to 3.3 times as many steps as the median global bound (Table \ref{tab:window}), although it is still conservative, and the horizon calibrated with the median one-step error predicted the observed saturation lead to within 25\% for the representative reservoirs (Table \ref{tab:horizon}); with the worst-case one-step error, the saturation lead for the Lorenz system exceeds $n_{\max}$ by a factor of about two. For the two maps, every admissible root of $G^n$ with $n\le4$ matched a true periodic point, with completeness verified by exact counts (Table \ref{tab:orbit-search} and the Appendix), and for the Lorenz system reservoir orbits of prime period 16 lay within $8\times10^{-4}$ of the shortest periodic orbit of the time-discretized flow, with leading multipliers accurate to approximately $0.1\%$, although the contracting multiplier was not resolved (Table \ref{tab:lorenz-orbits}). These recovery results are empirical, since the converse direction is not proved, and the theory is conditional on the orbits remaining in $K$ and on a bound $\varepsilon$ that is measured rather than predicted. The theorems do not invoke generalized synchronization \citep{lymburn2019}, as they depend on the reservoir only through the one-step errors; synchronization enters indirectly, by allowing the readout to approximate $f$ from the reservoir state (Section \ref{sec:gs-esp}). Further limitations are that prediction is reliable only for $n\lesssim n_{\max}$, that the unregularized readout loosens the bound of Corollary \ref{cor:periodic_detection}, and that each system was studied noise-free at a single parameter set. Extending the analysis across parameter values is a further direction. The error bounds of Theorems \ref{thm:error_propagation} and \ref{thm:local_jacobian} depend on the reservoir only through the one-step errors, so they are formally applicable to other readouts, such as the linearly combined readouts of \citet{ChenInubushi2026}, with $L$, $K$, and $\varepsilon$ taken for the target system at the corresponding parameter. Whether this yields useful bounds beyond the training parameters is left open. Proving the converse, bounding $\varepsilon_n$ in terms of the training data and the reservoir parameters alone, and extending the method to noisy and higher-dimensional systems are natural directions for future work.

\begin{acknowledgments}
	The authors gratefully acknowledge the financial support received through the University Research Committee (URC) scholarship at the University of Johannesburg, South Africa. The opinions and conclusions expressed in this manuscript are exclusively those of the authors and do not represent the views of the University of Johannesburg or the University of Pretoria.
\end{acknowledgments}

\section*{AUTHOR DECLARATIONS}

\subsection*{Conflict of Interest}
The authors have no conflicts to disclose.

\subsection*{Author Contributions}
\textbf{Taheer Jooma Abbajee:} Conceptualization; Methodology; Software; Validation; Formal analysis; Investigation; Data curation; Writing -- original draft; Writing -- review and editing; Visualization; Project administration.
\textbf{Keegan Doig Anderson:} Conceptualization; Supervision; Project administration; Validation; Writing -- review and editing.
\textbf{Muaaz Bhamjee:} Conceptualization; Supervision; Project administration; Validation; Writing -- review and editing.
\textbf{Maria Vivien Visaya:} Conceptualization; Supervision; Project administration; Validation; Writing -- review and editing.

\subsection*{AI Disclosure}
Claude (Anthropic) was used for limited technical and presentation assistance, including TikZ diagram preparation, figure styling suggestions, and debugging code and \LaTeX{} errors. The authors reviewed the resulting material and take full responsibility for the content of the manuscript.

\section*{DATA AVAILABILITY}
The data supporting the findings of this study were generated using code that is publicly available on \href{https://github.com/uj-cimru/autonomous_esn_bounds}{GitHub}.

\appendix*
\section{Experimental Details and Additional Tables}
\label{app:protocol}
\noindent\textit{Data and domains.} The domain $K$ is the bounding box of a 20{,}000-step trajectory after a 1{,}000-step burn-in, inflated by 5\% of each coordinate's range; $\operatorname{diam}(K)$ is the Euclidean norm of the box diagonal. To three significant figures, $K=[0.0414, 1.02]$ for the logistic map, $[-1.41, 1.40]\times[-0.424, 0.420]$ for H\'enon and $[-20.7, 20.6]\times[-27.1, 27.9]\times[0.0390, 47.7]$ for Lorenz. The reservoir is fed physical coordinates, except for the Lorenz system, for which normalized coordinates are used, defined by $z=(x-C)/S$ with $C=(0,0,24)$ and $S=(20,28,24)$. All predictions and subsequent analyses are performed in physical coordinates. The data orbit has length $T_1+2n_{\text{lead}}$ and is the same for all reservoir seeds. The reservoir is warmed up for $T_0$ steps and trained on the pairs $(r_t,x_{t+1})$, $t=T_0,\dots,T_1-1$. Exact reservoir hyperparameters are listed in Table \ref{tab:app-setup}. Each reservoir realization is generated from an integer seed between 0 and 49, which initializes NumPy's \texttt{default\_rng} for the random reservoir weights (the same 50 seeds are used for every system). The hyperparameters reported in Table \ref{tab:app-setup} were selected to minimize the training root mean square error using seed 50, which was reserved for hyperparameter selection and is not included among the 50 reservoir realizations (seeds 0 to 49) used for numerical validation.

\noindent\textit{Echo-state diagnostics.} The symmetric matrix $W_{\text{res}}$ is rescaled to spectral radius $\rho$, so $\norm{W_{\text{res}}}_2=\rho<1$ holds exactly for all 150 reservoirs. The echo state property is certified by the negative largest conditional Lyapunov exponent $\lambda_{\text{cond}}$, estimated from a single renormalized tangent vector over a 3{,}000-step drive after a 500-step transient. The estimated exponents are $-0.643$, $-0.642$ and $-0.379$ for the logistic, H\'enon and Lorenz reservoirs, respectively, confirming that the ESP criterion $\lambda_{\text{cond}} < 0$ is satisfied in all three cases.

\noindent\textit{Root finding and matching.}
All root-finding uses a multi-start search \citep{auerbach1987} with the Powell hybrid method implemented in \texttt{scipy.\allowbreak optimize.\allowbreak root} \citep{virtanen2020}, with tolerance $10^{-10}$ on $\norm{f^n(x)-x}$ for true periodic points and $10^{-6}$ on $\norm{G^n(r)-r}$ for reservoir periodic orbits, checked explicitly because the solver's own flag only tests the step size. Roots are deduplicated at $10^{-7}\operatorname{diam}(K)$ (true points) and $10^{-4}$ in reservoir space (reservoir roots). One-to-one matching uses the Hungarian algorithm \citep{kuhn1955}, minimizing total Euclidean distance (\texttt{scipy.\allowbreak optimize.\allowbreak linear\_\allowbreak sum\_\allowbreak assignment}).

\noindent\textit{Tables.} Table \ref{tab:app-setup} gives the reservoir hyperparameters at full precision. Table \ref{tab:summary} summarizes every check, and Table \ref{tab:control} the negative control. Table \ref{tab:app-periodic} lists the true periodic points by order, including the exact counts used to verify completeness. Table \ref{tab:app-leads} gives per-lead medians for the error-propagation experiment, and Table \ref{tab:app-seedsweep} the 50-seed sweep, in which the full battery of Section \ref{sec:val-exp1} was run for every seed.

\begin{table*}
	\centering
	\caption{Reservoir hyperparameters, representative seed and data protocol.}
	\label{tab:app-setup}
	{\footnotesize\setlength{\tabcolsep}{4pt}
		\begin{tabular}{@{}lccc@{}}
			\toprule
			& Logistic ($r{=}3.91$) & H\'enon ($a{=}1.4,\,b{=}0.3$) & Lorenz ($h{=}0.0974$) \\
			\midrule
			$d_r$ & 30 & 50 & 100 \\
			$\rho$ & $0.9491003072336582$ & $0.9393444640378177$ & $0.9429962701935458$ \\
			$\tau$ & $0.987$ & $0.9971$ & $0.998$ \\
			$\sigma$ & $2.291716793108812$ & $0.03972440210357811$ & $1.1137173954268642$ \\
			$\beta$ & $9.82285158135535\times10^{-5}$ & $0.7082107864030838$ & $0.07154312197661555$ \\
			seed & 15 & 37 & 1 \\
			$T_0$ & 50 & 50 & 50 \\
			$T_1$ & 100 & 150 & 350 \\
			$n_{\text{lead}}$ & 70 & 110 & 230 \\
			$x_0$ & $0.63$ & $(0,0)$ & $(1.0, 1.4, 28.8)$ \\
			\bottomrule
		\end{tabular}
	}
\end{table*}

\begin{table*}
	\centering
	\caption{Summary of all checks. Admissible: admissible cases graded. Violations: cases in which a proven bound was violated. Non-vacuous: individual (case, bound) checks with the bound below $\operatorname{diam}(K)$, out of five bounds per case for the error-propagation experiment and for true periodic points, and four per case for the fixed-point residuals. The negative control is reported in Table \ref{tab:control}.}
	\label{tab:summary}
	{\small
		\begin{tabular}{@{}llrrr@{}}
			\toprule
			Experiment & Statement & admissible & violations & non-vacuous \\
			\midrule
			Section \ref{sec:val-exp1} & five bounds, representative seeds & 69\,900 & 0 & 35\,780 \\
			& five bounds, 50 seeds per system & 3\,486\,549 & 0 & n/a \\
			Section \ref{sec:val-exp2} & residual at true periodic points & 52 & 0 & 212 \\
			& residual growth from fixed points & 750 & 0 & 454 \\
			Section \ref{sec:val-exp3} & roots of $G^n$, all bounds & 61 & 0 & n/a \\
			\bottomrule
		\end{tabular}
	}
\end{table*}

\begin{table*}
	\centering
	\caption{Negative control (representative seeds): number (fraction) of admissible (origin, lead) cases in which the effective bound is violated when $L$ and $\ell_k$ are replaced by $s\,L$ and $s\,\ell_k$. Overall, 127\,317 of 139\,800 checks (91.1\%) fail.}
	\label{tab:control}
	{\small
		\begin{tabular}{@{}lrrrr@{}}
			\toprule
			System & $s$ & cases & global fails & local fails \\
			\midrule
			Logistic & 0.3 & 4\,900 & 3\,851 (78.6\%) & 4\,617 (94.2\%) \\
			H\'enon & 0.3 & 12\,100 & 7\,286 (60.2\%) & 11\,596 (95.8\%) \\
			Lorenz & 0.1 & 52\,900 & 49\,077 (92.8\%) & 50\,890 (96.2\%) \\
			\bottomrule
		\end{tabular}
	}
\end{table*}

\begin{table*}
	\centering
	\caption{True points of period $n$ in $K$ (representative seeds), that is, points with $f^n(x)=x$. Found: points located by Newton search. Exact: independent exact count of the real fixed points of $f^{n}$ in $K$ (for the logistic map excluding $x=0\notin K$), and the analytic equilibria for Lorenz $n=1$. Prime period $n$: the found points whose prime period is exactly $n$; the others have a smaller prime period dividing $n$. Max residual: largest $\norm{\fh{n}(x^*)-x^*}$ over the found points, in physical units. For Lorenz $n=16$ the 16 points form one orbit of prime period 16 on the closed curve contained in $\Fix(f^{16})$; the two equilibria are counted at $n=1$.}
	\label{tab:app-periodic}
	{\small
		\begin{tabular}{@{}llrrrr@{}}
			\toprule
			System & $n$ & found & exact & prime period $n$ & max residual \\
			\midrule
			Logistic & 1 & 1 & 1 & 1 & $3.20\times10^{-10}$ \\
			& 2 & 3 & 3 & 2 & $1.82\times10^{-8}$ \\
			& 3 & 7 & 7 & 6 & $1.45\times10^{-8}$ \\
			& 4 & 7 & 7 & 4 & $2.24\times10^{-7}$ \\
			\addlinespace
			H\'enon & 1 & 2 & 2 & 2 & $4.90\times10^{-8}$ \\
			& 2 & 4 & 4 & 2 & $2.05\times10^{-7}$ \\
			& 3 & 2 & 2 & 0 & $7.15\times10^{-7}$ \\
			& 4 & 8 & 8 & 4 & $2.37\times10^{-6}$ \\
			\addlinespace
			Lorenz & 1 & 2 & 2 & 2 & $9.84\times10^{-3}$ \\
			& 16 & 16 & n/a & 16 & $6.91\times10^{-3}$ \\
			\bottomrule
		\end{tabular}
	}
\end{table*}

\FloatBarrier
\begin{table*}
\centering
\caption{Medians over admissible origins at selected leads (representative seeds): iterated error $E_n$, effective local and global bounds, and the uniform bound $B_L(n)\varepsilon_n(\mathcal P)$. At $n=1$ the effective local and global bounds equal the one-step error of each origin, so their medians equal the median $E_1$; the uniform bound equals the supremum $\varepsilon_1(\mathcal P)$ over origins and is therefore larger.}
\label{tab:app-leads}
{\small
\begin{tabular}{@{}lrrrrr@{}}
\toprule
System & $n$ & median $E_n$ & effective local & effective global & uniform\\
\midrule
Logistic & 1 & $1.0\times10^{-9}$ & $1.0\times10^{-9}$ & $1.0\times10^{-9}$ & $1.2\times10^{-8}$ \\
		& 5 & $8.6\times10^{-8}$ & $6.1\times10^{-7}$ & $1.1\times10^{-5}$ & $4.5\times10^{-5}$ \\
		& 10 & $1.5\times10^{-6}$ & $2.4\times10^{-5}$ & $0.0189$ & $0.0604$ \\
		& 20 & $3.9\times10^{-4}$ & $7.3\times10^{-3}$ & $3.0\times10^{4}$ & $7.8\times10^{4}$ \\
		& 30 & $0.0629$ & $4.14$ & $4.4\times10^{10}$ & $1.0\times10^{11}$ \\
		& 40 & $0.26$ & $5.0\times10^{4}$ & $6.5\times10^{16}$ & $1.3\times10^{17}$ \\
		\addlinespace
		H\'enon & 1 & $2.1\times10^{-10}$ & $2.1\times10^{-10}$ & $2.1\times10^{-10}$ & $6.7\times10^{-9}$ \\
		& 5 & $1.9\times10^{-9}$ & $2.2\times10^{-8}$ & $2.7\times10^{-7}$ & $2.5\times10^{-6}$ \\
		& 10 & $1.3\times10^{-8}$ & $1.2\times10^{-6}$ & $4.1\times10^{-4}$ & $4.2\times10^{-3}$ \\
		& 20 & $9.5\times10^{-7}$ & $2.2\times10^{-3}$ & $781$ & $5.5\times10^{3}$ \\
		& 30 & $4.9\times10^{-5}$ & $2.98$ & $1.2\times10^{9}$ & $7.3\times10^{9}$ \\
		& 40 & $3.9\times10^{-3}$ & $5.0\times10^{3}$ & $1.6\times10^{15}$ & $9.6\times10^{15}$ \\
		& 50 & $0.127$ & $1.1\times10^{7}$ & $2.1\times10^{21}$ & $2.1\times10^{22}$ \\
		\addlinespace
		Lorenz & 1 & $2.2\times10^{-4}$ & $2.2\times10^{-4}$ & $2.2\times10^{-4}$ & $0.0464$ \\
		& 5 & $8.6\times10^{-4}$ & $8.4\times10^{-3}$ & $0.418$ & $34.9$ \\
		& 10 & $2.9\times10^{-3}$ & $0.0982$ & $1.4\times10^{3}$ & $9.9\times10^{4}$ \\
		& 20 & $8.2\times10^{-3}$ & $6.34$ & $1.8\times10^{10}$ & $8.2\times10^{11}$ \\
		& 30 & $0.0219$ & $443$ & $2.4\times10^{17}$ & $6.6\times10^{18}$ \\
		& 50 & $0.127$ & $2.7\times10^{6}$ & $6.2\times10^{31}$ & $5.8\times10^{32}$ \\
		& 100 & $3.74$ & $6.5\times10^{15}$ & $6.0\times10^{66}$ & $2.8\times10^{67}$ \\
	\bottomrule
\end{tabular}
}
\end{table*}
\FloatBarrier

\begin{table*}
	\centering
	\caption{Robustness across 50 reservoir realizations per system: minimum, median and maximum of the one-step error $\varepsilon_1$, the horizon $n_{\max}$, the observed saturation lead $n_{\mathrm{emp}}$, and the ratio $n_{\mathrm{emp}}/n_{\max}$. No bound or ordering violation occurs in any of the 245\,000 (logistic), 605\,000 (H\'enon) and 2\,636\,549 (Lorenz, 99.7\% of $50\times230\times230$) admissible cases, and no reservoir has $\norm{W_{\text{res}}}_2\ge1$.}
	\label{tab:app-seedsweep}
	{\small
		\begin{tabular}{@{}llrrrr@{}}
			\toprule
			System & & $\varepsilon_1$ & $n_{\max}$ & $n_{\mathrm{emp}}$ & $n_{\mathrm{emp}}/n_{\max}$ \\
			\midrule
			Logistic & min & $5.46\times10^{-10}$ & 30.8 & 19 & 0.51 \\
			& median & $9.65\times10^{-9}$ & 36.6 & 29 & 0.77 \\
			& max & $1.68\times10^{-7}$ & 42.5 & 32 & 0.88 \\
			\addlinespace
			H\'enon & min & $1.06\times10^{-9}$ & 42.2 & 50 & 1.07 \\
			& median & $9.54\times10^{-9}$ & 44.7 & 52 & 1.17 \\
			& max & $2.66\times10^{-8}$ & 49.8 & 55 & 1.23 \\
			\addlinespace
			Lorenz & min & $2.22\times10^{-2}$ & 41.2 & 42 & 0.83 \\
			& median & $6.69\times10^{-2}$ & 53.9 & 93.5 & 1.73 \\
			& max & $2.09\times10^{-1}$ & 66.3 & 115 & 2.16 \\
			\bottomrule
		\end{tabular}
	}
\end{table*}

\clearpage
\bibliography{refs.bib}

\end{document}